\documentclass[lettersize,journal]{IEEEtran}
\usepackage{amsmath,amssymb,amsfonts}
\usepackage{algorithmicx,algorithm}
\usepackage{amsthm,amsmath,amssymb}
\usepackage{mathrsfs}
\usepackage{array}
\usepackage{textcomp}
\usepackage{stfloats}
\usepackage{verbatim}
\usepackage{graphics} 
\usepackage{epsfig} 
\usepackage{times} 
\usepackage{amsmath} 
\usepackage{amssymb}  
\usepackage{cite} 
\usepackage{graphicx}
\usepackage{subfigure} 
\usepackage{picinpar}
\usepackage{url}
\usepackage{flushend}
\usepackage{colortbl}
\usepackage{soul}
\usepackage{multirow}
\usepackage{pifont}
\usepackage{color}
\usepackage{alltt}
\usepackage{hyperref}
\usepackage{float}
\usepackage{enumerate}
\usepackage{siunitx}
\usepackage{breakurl}
\usepackage{epstopdf}
\usepackage{pbox}
\usepackage{booktabs}
\usepackage{makecell} 
\usepackage{threeparttable} 
\usepackage{balance}
\usepackage[justification=centering,labelsep=period]{caption} 
\usepackage{color} 
\usepackage{gensymb} 
\usepackage{upgreek}
\hypersetup{colorlinks=true, linkcolor=black} 

\usepackage{adjustbox}
\usepackage{xcolor}
\usepackage{soul}
\soulregister\cite7 
\soulregister\citep7 
\soulregister\citet7 
\soulregister\ref7 
\soulregister\pageref7 

\usepackage{bm}
\usepackage{geometry}
\title{\LARGE \bf
PL-EVIO: Robust Monocular Event-based Visual Inertial Odometry with Point and Line Features 
}

\author{Weipeng Guan,   Peiyu Chen,   Yuhan Xie,   Peng Lu$^{*}$
\thanks{
This work was supported by General Research Fund under Grant 17204222, and in part by the Seed Fund for Collaborative Research and General Funding Scheme-HKU-TCL Joint Research Center for Artificial Intelligence.

The authors are with the Adaptive Robotic Controls Lab (ArcLab), Department of Mechanical Engineering, Faculty of Engineering, The University of Hong Kong, Hong Kong SAR, China.
(Weipeng Guan and Peiyu Chen contributed equally to this work, corresponding author: lupeng@hku.hk). 

Demonstrations: \url{https://b23.tv/pKVbvPJ}
}%
}

\begin{document}   
\maketitle  
\thispagestyle{plain} 
\pagestyle{plain} 


\begin{abstract}
Robust state estimation in challenge situations is still an unsolved problem, especially achieving onboard pose feedback control for aggressive motion.
In this paper, we propose robust and real-time event-based visual-inertial odometry (VIO) that incorporates event, image, and inertial measurements. 
Our approach utilizes line-based event features to provide additional structure and constraint information in human-made scenes, while point-based event and image features complement each other through well-designed feature management.
To achieve reliable state estimation, we tightly couple the point-based and line-based visual residuals from the event camera, the point-based visual residual from the standard camera, and the residual from IMU pre-integration using a keyframe-based graph optimization framework.
Experiments in the public benchmark datasets show that our method can achieve superior performance compared with the state-of-the-art image-based or event-based VIO.
Furthermore, we demonstrate the effectiveness of our pipeline through onboard closed-loop quadrotor aggressive flight and large-scale outdoor experiments.
Videos of the evaluations can be found on our website: \url{https://youtu.be/KnWZ4anBMK4}.

\end{abstract}

\vspace{0.5em}
\def\abstractname{Note to Practitioners}
\begin{abstract}

Driven by the need for real-time closed-loop control for drones under aggressive motion and broad illumination environments, many existing VIO systems fail to meet these requirements due to the inherent limitations of standard cameras.
Event cameras are bio-inspired sensors that capture pixel-level illumination changes instead of the intensity image with a fixed frame rate, which can provide reliable visual perception during high-speed motions and in high dynamic range scenarios.
Therefore, developing state estimation algorithms based on event cameras offers exciting opportunities for robotics.
However, adopting event cameras is challenging due to the event streams being composed of asynchronous events which are fundamentally different from the synchronous intensity images.
Moreover, event cameras output minimal
information or even noise when the relative motion between the camera and the scene is limited, such as in a still state, while standard cameras can provide rich perception information in most scenarios.
In this paper, we propose a robust, high-accurate, and real-time optimization-based monocular event-based VIO framework that tightly fuses the event, image, and IMU measurement together.
Owing to the well-designed framework and good feature management, our system can provide robust and reliable state estimation in challenging environments. 
The efficiency of our system is adequate to achieve real-time operation on platforms with limited resources, such as providing onboard pose feedback for quadrotor flights.

\end{abstract}


\section{INTRODUCTION}
\label{INTRODUCTION}
\subsection{Motivations} 
\label{Motivations}

\begin{figure}[htb]  
        \vspace{-1.0em}
        \setlength{\abovecaptionskip}{-0.01em}
    	\subfigtopskip=0pt 
    	\subfigbottomskip=-1pt 
    	\subfigcapskip=-6pt 
        \centering
        \captionsetup{justification=justified}
        \subfigure[ ]{
                \begin{minipage}[t]{1.0\columnwidth}
                \centering
                \includegraphics[width=1.0\columnwidth]{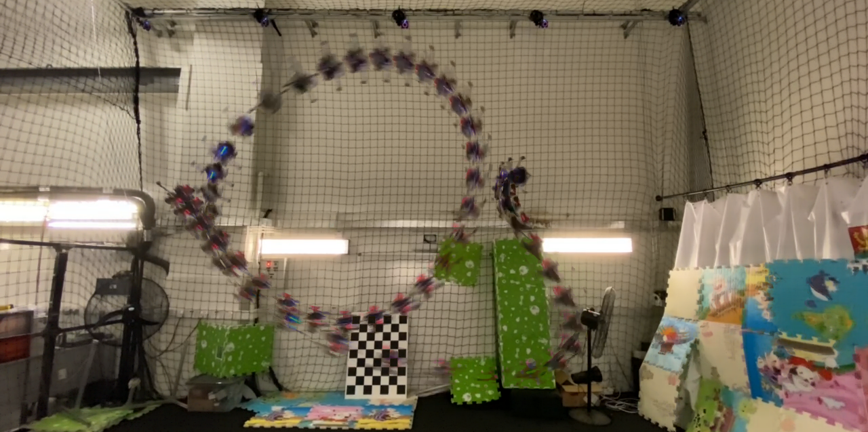}
                \label{first_image_a}
                \end{minipage}%
        }
        \subfigure[ ]{
                \begin{minipage}[t]{1.0\columnwidth}
                \centering
                \includegraphics[width=1.0\columnwidth]{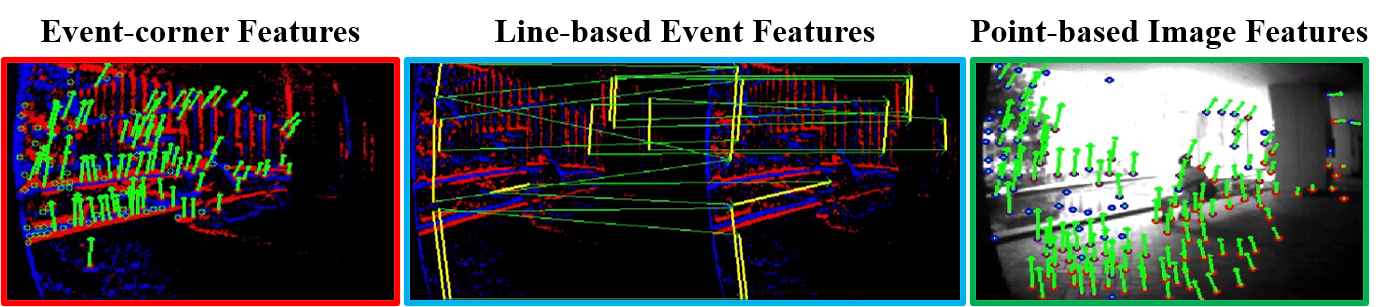}
                \label{first_image_b}
                \end{minipage}%
        }
        \caption{
        (a) Our PL-EVIO combines events, images, and IMU to provide robust state estimation during aggressive motion.
        It can provide onboard feedback-control for quadrotors with limited computational resources. 
        (b) Our PL-EVIO in the outdoor environment.
            Left: event-corner features in the event; 
            Middle: line-based features in the event; 
            Right: point-based features in the image.
        }  
        \label{first_image}
        \vspace{-1.0em}
\end{figure}

\IEEEPARstart{S}{tandard}
cameras have inherent limitations, including sensing latency and low dynamic range, which is challenging for image-based Visual Odometry (VO), Visual Inertial Odometry (VIO), and Simultaneous Localization and Mapping (SLAM) systems to detect and track features under high-speed motion or high-dynamic-range (HDR) scenarios.
Specifically, robust state estimation is vital for real-time feedback control in aggressive motion (e.g. onboard quadrotor flip, as shown in Fig.\ref{first_image_a}), since even tiny drifts or momentary poor feature tracking can potentially lead to a crash.
Event cameras offer exciting opportunities to solve the aforementioned problems, which possess several advantages over standard cameras, including low latency ($\upmu$s-level), HDR (140 dB), and no motion blur \cite{GWPHKU:EVENT-SURVEY}.

Most of the research in both image and event-based SLAM/VO/VIO rely on point-based features, while it is important to note that human-made structures often exhibit regular geometric shapes, such as lines or planes. 
Therefore, point-based features may not always be the optimal representation for visual tracking in all scenarios.
Performance degeneracy might occur when only using point-based features, while point-based features are more common in natural scenes.
Therefore, for heterogeneous event-based information utilization, we design and extract the line-based feature in the event stream to improve the performance of purely point-based features, since the line-based features can reflect more geometric structure information than point-based features \cite{GWPHKU:PL-SLAM} \cite{xu2022leveraging} \cite{GWPHKU:PL-VINS}.
As can be seen in Fig.\ref{first_image_b} Fig.\ref{Point_Line_feature}, and Fig.\ref{Human-made-scenarios}, the integration of the line-based feature and point-based feature can further ensure a more uniform distribution of the features and provide additional constraints on scene structure.

In addition, compared to standard cameras, event cameras are capable of providing reliable visual perception during high-speed motion and HDR scenarios.
However, when the event camera and the scene have restricted relative motion, such as in a static state, event cameras may produce limited information or even noise.
Although the standard camera encounter difficulties during high-speed motion or in HDR scenarios, it can provide rich intensity value of the scenes under uniform motion or favorable lighting conditions.

Observing this complementary, we propose a monocular VIO framework for a sensor setup that includes event, image, and inertial measurement unit (IMU) data, with a well-designed feature management system.
Our VIO framework includes the purely event-based VIO (EIO), and the event with image-based VIO (EVIO).
More specifically, we first implement a motion compensation algorithm using the IMU data to correct the motion of each event according to its individual timestamp, including rotation and translation motion, into the same timestamp.
After that, utilizing the event-corner features detection and tracking approach developed in our previous EIO work\cite{GWPHKU:MyEVIO}. We conduct an EIO framework, including the line-based event features and the event-corner point features, termed PL-EIO (Event+IMU), to perform robust state estimation.
Finally, we integrate image measurements into our PL-EIO framework as the PL-EVIO (Event+Image+IMU), in which visual landmarks include event-corner features, line-based event features, and point-based image features.
These three kinds of features are well integrated together to leverage additional structure or constraint information for more accurate and robust state estimation. 

\subsection{Contributions} 
\label{Contributions}
Our contributions are summarized as follows:

\begin{enumerate}
\item In order to handle the HDR situations and aggressive motion, especially the onboard aggressive motion, we propose the PL-EVIO pipeline, which tightly fuses the event-corner features, line-based event features, and point-based image features together, to provide robust and reliable state estimation.
\item To address the performance degradation when only using point-based features in human-made structures, we design the line-based feature and descriptor in event-based representation for front-end incremental estimation.
\item We validate that our PL-EVIO can achieve state-of-the-art performance in different challenging datasets. It also can be used as onboard pose feedback control for the quadrotor to achieve aggressive motion, e.g. flip.


\end{enumerate}

The remainder of the paper is organized as follows:
Section \ref{Related Works} introduces the related works. 
Section \ref{Methodology} introduces the principle of our proposed method.
Section \ref{Evaluation} presents the experiments and results.
Finally, the conclusion is given in Section \ref{CONCLUSIONS}.

\section{Related Works}
\label{Related Works}
\subsection{Event-based Representation and Feature Extraction} 
\label{Event-based Representation and Feature Extraction}
Event cameras are motion-activated sensors that capture pixel-level illumination changes instead of the intensity image with a fixed frame rate. 
An event is triggered only when the intensity of an individual pixel varies beyond a specific threshold $T_{threshold}$, which can be represented as the spatio-temporal coordinates of the intensity change and its sign: 
\begin{equation}
e=\left\{t,x,y,p \right\} \Leftrightarrow I(x, y, t+\bigtriangleup t)-I(x, y, t)=p \cdot T_{threshold}
\end{equation}
where $ t $ is the timestamp that the intensity of a pixel $ I\left( x,y \right)$ changes, and $ p $ is the polarity that indicates the direction of the intensity change.
The generation model of the event stream endowed some good properties, which also allow the event camera to confer robustness to vision-based localization in challenging scenarios. 
However, adopting the event camera into the SLAM/VO/VIO is a very challenging task since the event streams are in asynchronous formats which is fundamentally different from the synchronous image data. Therefore, most methods and concepts developed for conventional image-based cameras can not be directly applied.
To enable the asynchronous event data into the synchronous data representation, different kinds of event representation have been proposed: 

$(i)$ The first method is directly working on the raw event stream without any frame-like accumulation. 
Ref.\cite{GWPHKU:ACE} proposed a feature tracker that employe the descriptors for event data.
Ref.\cite{GWPHKU:zhu2017event} presented a feature tracker based on Expectation Maximisation (EM).
Ref.\cite{GWPHKU:PowerLine} \cite{GWPHKU:IDOL} extracted the line feature from the raw asynchronous events.
There are several other ways to represent the raw event, such as Voxel Grid or Event spike tensor\cite{chen2020event}. 
However, these higher dimensional or learning-based event representations will not be discussed here.

$(ii)$ The second approach is combining with the image sensor, or generating the intensity image from the event through learning-based methods.
Ref.\cite{GWPHKU:kueng2016low} \cite{GWPHKU:EKLT} firstly detected the features on the grayscale image frames, and then track the features asynchronously using event streams. 

$(iii)$ The third representation is the motion-compensated event image, or edge image, which is generated by aggregating a group of neighbor events within the spatio-temporal window into an edge image. 
Ref. \cite{GWPHKU:ETH-EVIO} \cite{GWPHKU:Ultimate-SLAM} adopted the conventional corner detection algorithms, such as FAST corners \cite{fast_corner_detection} or Shi-Tomasi \cite{Shi-Tomasi} for feature detection, and the Lucas Kanade (LK) optical flow \cite{LK_optical_flow} for feature tracking in the event image.

$(iv)$ The last method is the time surface (TS) or Surface of Active Event (SAE), which is a 2D map where each pixel stores the time value. 
It can summarize and update the event stream at any given instant, or encode the spatio-temporal constraints of the historical events. 
Using an exponential decay kernel, TS can emphasize recent events over past events\cite{GWPHKU:TS}. 
$t_{last}$ is the timestamp of the last event at each pixel coordinate $\boldsymbol{x}=(u,v)^{T}$, the TS at time $t \geq t_{last}(\boldsymbol{x})$ is defined by:
\vspace{-0.15cm}
\begin{equation}
T(\boldsymbol{x},t)=\exp(-\frac{t-t_{last}(\boldsymbol{x})}{\eta}) 
\label{TS_p}
\end{equation}
where $\eta$ is the decay rate parameter.
Ref.\cite{GWPHKU:Harris-event-corner} \cite{GWPHKU:ARC*} use the SAE or TS to inspect previously triggered events in the stream and the adjacent pixels for classifying a new event as an event-corner.

\subsection{Event-based Motion Estimation} 
\label{Event-based Motion Estimation}
Event-based state estimation has been extensively developed to handle challenging scenes in recent years, particularly in scenarios where traditional cameras struggle to perform well, such as high-speed motion estimation or HDR perception.
Ref. \cite{GWPHKU:first-event-SLAM} proposed the first event-based SLAM system, which is limited to tracking planar motions while reconstructing the 2D ceiling map with an upward-looking event camera. 
Ref.\cite{GWPHKU:censi2014low} and \cite{GWPHKU:kueng2016low} proposed the event-based VO to track camera motion. However, these methods still relied on the standard camera, which was still susceptible to motion blur and low dynamic range.
The first purely event-based 6-DoF (Degree-of-Freedom) VO was presented in \cite{GWPHKU:kim2016real}, which performed real-time event-based SLAM through three decoupled probabilistic filters that jointly estimate the 6-DoF camera pose, 3D map of the scene, and image intensity. However, it is computationally expensive and requires GPU to achieve real-time performance.
EVO \cite{GWPHKU:EVO} was proposed to solve the SLAM problem without recovering image intensity, thus reducing computational complexity, and it can run in real-time on a standard CPU. It performs a tracking approach based on image-to-model alignment and adopts the 3D reconstruction method from EMVS \cite{GWPHKU:EMVS} to perform the mapping. However, the EVO is needed to run in the scene that is planar to the sensor, for up to several seconds, for bootstrapping the system.  
ESVO \cite{GWPHKU:ESVO} is the first stereo event-based VO method, which follows a parallel tracking-and-mapping scheme to estimate the ego-motion and the semi-dense 3D map of the scene. However, it barely operates in real-time in DAVIS346 (346*260) and also faces limitations due to rigorous initialization as well as unreliable pose tracking.
Ref. \cite{GWPHKU:Feature-based-ESVO} proposed stereo VO for event cameras based on features. The pose estimation is done by re-projection error minimization, while the features are stereo and temporally matched through the consecutive left and right event TS. 
It solves the problems of ESVO mentioned above.
However, it still cannot operate in real-time in high-resolution event cameras (640*480).

\begin{figure*}[htb]  
        \centering
        \includegraphics[width=2.0\columnwidth]{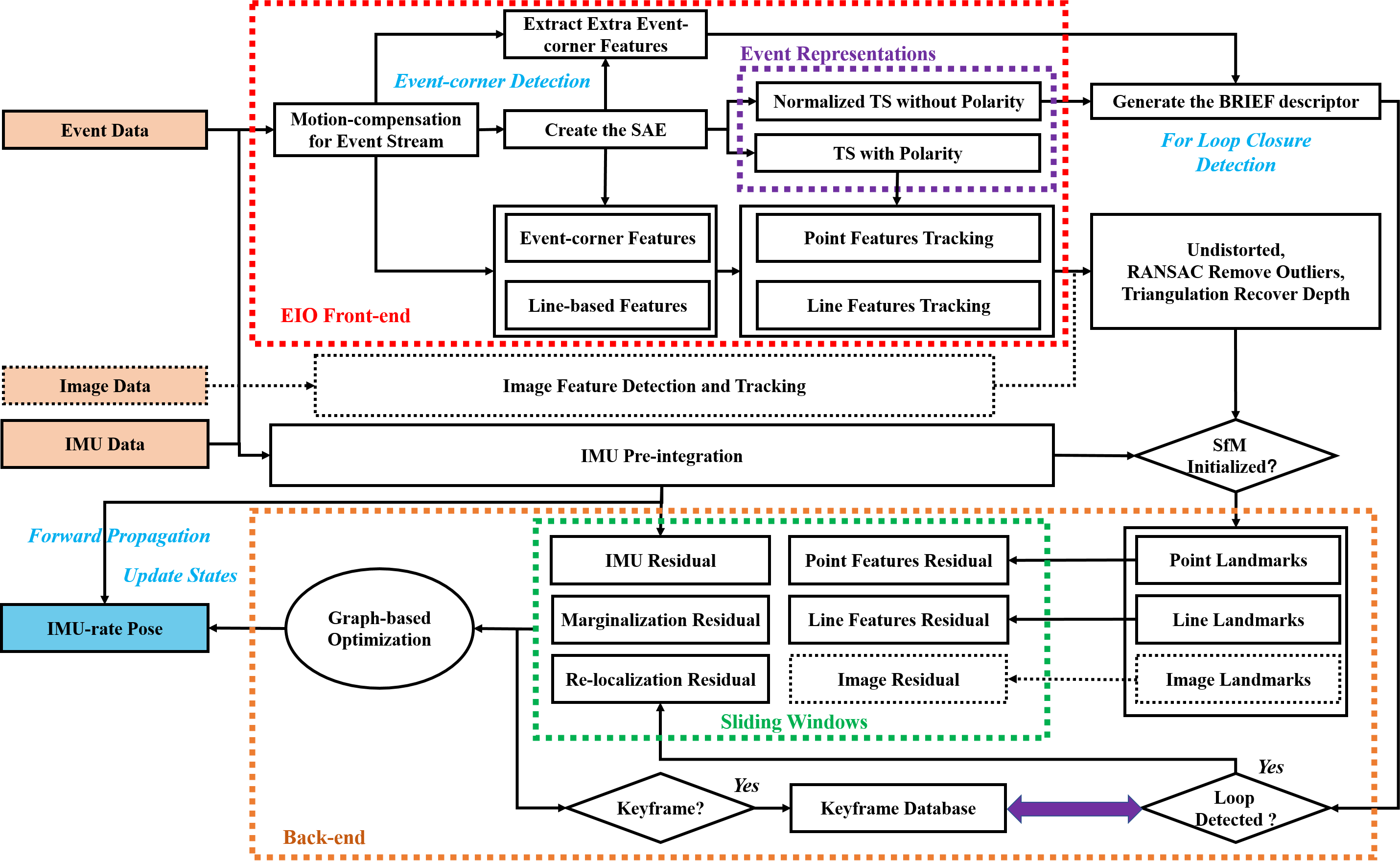}
        \caption{The framework of our PL-EIO (Event+IMU) and PL-EVIO (Event+Image+IMU)}  
        \label{pl-eio_framework}
        \vspace{-1.0em}
\end{figure*}%

The robustness of event-based SLAM/VO systems can be improved by incorporating IMU measurements.
The first EIO method was proposed in Ref. \cite{GWPHKU:Event-based-visual-inertial-odometry} which fused a purely event-based tracking algorithm with pre-integration IMU measurement through the Extended Kalman Filter. 
Another EIO method was proposed in Ref. \cite{GWPHKU:ETH-EVIO}. It detects and tracks the features in the edge image, which is generated from motion-compensated event streams, through traditional image-based feature detection and tracking methods. Finally, the tracked features are combined with IMU measurement using keyframe-based nonlinear optimization. 
The authors extended their method to leverage the complementary advantages of both standard and event cameras in Ultimate-SLAM \cite{GWPHKU:Ultimate-SLAM} to fuse events image frames, standard frames, and IMU. 
To some extent, these methods use the edge image to realize VIO, this might introduce bottlenecks since it requires substantial parameter adjustments depending on the varying number of generated events in the scene.
EKLT-VIO \cite{EKLT-VIO} combined the event-based tracker \cite{GWPHKU:EKLT} as the front-end with a filter-based back end to perform the EVIO for Mars-like sequences. However, it is pretty hard to perform in real-time even in the lowest resolution event camera.
Ref. \cite{GWPHKU:Continuous-time-visual-inertial-odometry-for-event-cameras} proposed to fuse events and IMU measurement into a continuous-time framework. 
While their approach cannot achieve real-time since the expensive optimization is required to update the spline parameters upon receiving every event \cite{GWPHKU:ETH-EVIO}.
In our previous work \cite{GWPHKU:MyEVIO}, we proposed a monocular EIO which the event-corner features with IMU measurement to provide real-time 6-DoF state estimation even in high-resolution event cameras. 
Furthermore, this EIO framework can bootstrap from unknown initial states and can ensure global consistency thanks to the loop closure function.
Nonetheless, it still cannot provide onboard pose estimation for closed-loop control of the quadrotor flight since event cameras produce minimal information or noise when stationary.
Recently, there have been several studies focusing on stereo EVIO\cite{GWPHKU:ESVIO}\cite{esvioliu2023}.

There are several works in event-based vision that utilize line features.
IDOL\cite{GWPHKU:IDOL} calculates the normal vectors in the spatio-temporal space for each incoming event by utilizing a local neighborhood.
Events with similar normal vectors are clustered together to form lines, and an EIO algorithm uses these detected lines and inertial measurements to estimate camera poses. 
However, this approach assumes that lines move at nearly a constant speed in short intervals, leading to the aggressive motion being avoided in their validation experiments and loss of the advantages of event cameras.
What's more, this method lacks real-time capabilities even with low-resolution event cameras (240*180).
Ref.\cite{chamorro2022event} employed the Hough transformation on spatial images generated from a 3D point-based map to cluster event data into a collection of 3D lines.
These lines are subsequently integrated into the Kalman filter to estimate the 3D lines and camera pose. 
However, their event-to-line matching method suffers from the sudden surge of incoming events\cite{chamorro2020high} caused by aggressive motion, scene complexity, and sudden illumination changes.
Additionally, this approach is sensitive to event sparsity and requires at least 6 non-parallel 3D lines, a known-scale predefined marker, or ground-truth pose readings for system bootstrapping.
Both spatio-temporal relationship\cite{GWPHKU:IDOL} and Hough transformation\cite{chamorro2022event} are utilized in event data to cluster events that belong to the same straight lines.
In contrast, our approach utilizes the line segment detector (LSD)\cite{GWPHKU:LSD} to extract event-based line features and strike a balance between performance and computational efficiency.
Unlike these two methods that solely rely on the line feature and may be susceptible to high levels of texture in the scene, our method leverages the complementarity of point and line features to enhance its robustness.
Moreover, Ref.\cite{GWPHKU:PowerLine} utilized event cameras for powerline tracking.
Their method involved detecting planes in the spatio-temporal signal to identify lines in the event streams and subsequently incorporating events into these lines while tracking them over time.
However, their approach is restricted to powerline inspection tasks and does not involve the data association of event-based line features or utilization for incremental pose estimation.

\section{Methodology} 
\label{Methodology}
\begin{figure*}[htb]  
        \setlength{\abovecaptionskip}{-0.1em}
    	\subfigtopskip=0pt 
    	\subfigbottomskip=0pt 
    	\subfigcapskip=-8pt 
        \centering
        \subfigure[]{
                \begin{minipage}[t]{\columnwidth}
                \centering
                \includegraphics[width=\columnwidth]{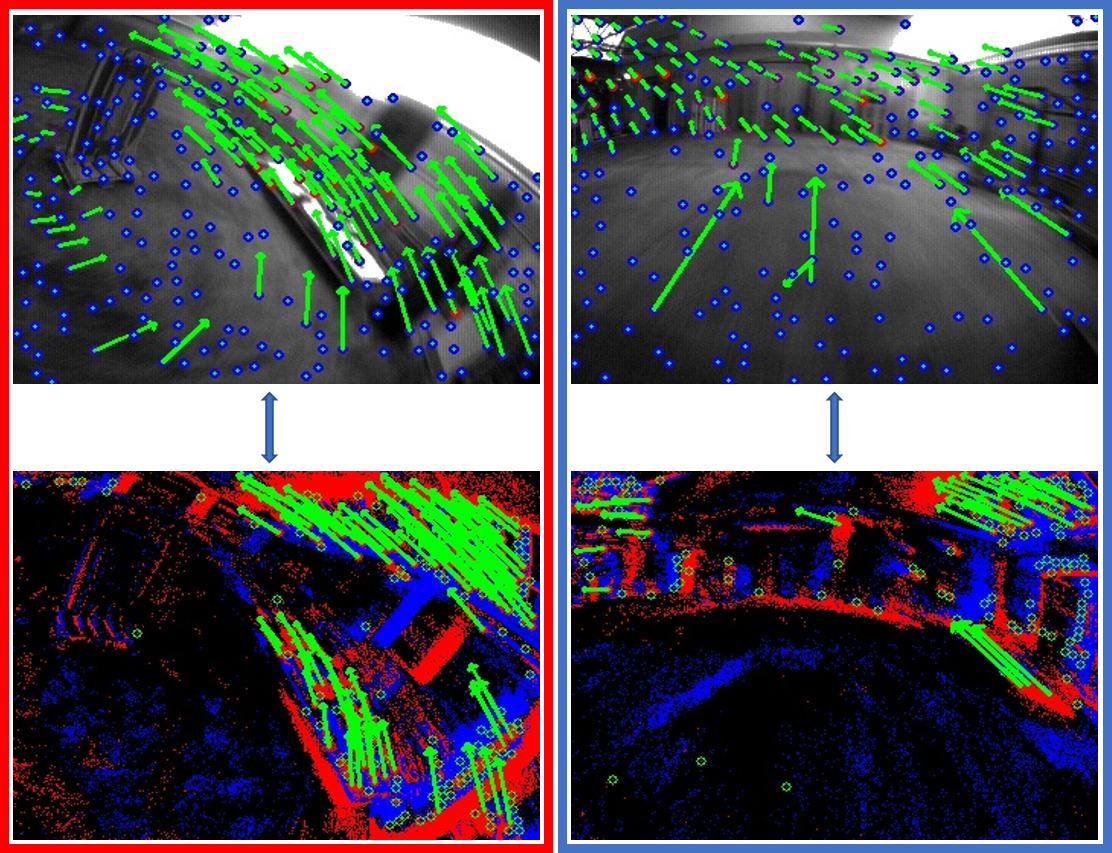}
                \label{Point_feature}
                \end{minipage}%
        }%
        \subfigure[]{
                \begin{minipage}[t]{\columnwidth}
                \centering
                \includegraphics[width=\columnwidth]{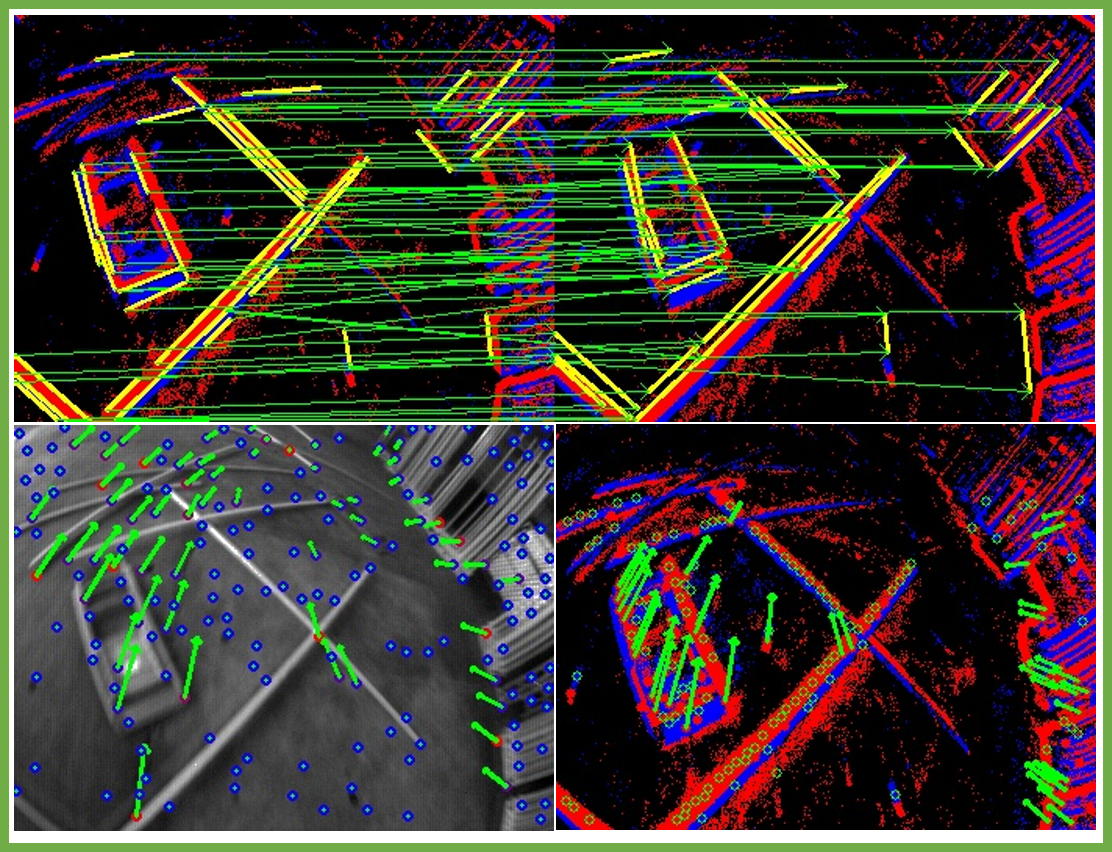}
                \label{Line_feature}
                \end{minipage}%
        }%
        \caption{Three different kinds of features in our PL-EVIO framework: event-corner features, event-based line features, and image-based point features}
        \label{Point_Line_feature}
        \vspace{-2.0em}
\end{figure*}%

\subsection{Framework Overview} 
\label{Framework Overview}
The structure of our proposed method is illustrated in Fig.\ref{pl-eio_framework}, which is composed of two sections: 
$ \left( i \right)$ The EIO Front-end takes the motion-compensated event stream as input and extracts the event-corner features and the line-based event features.
There are two kinds of event representations: the TS with polarity 
$T_{p}(\boldsymbol{x},t)=p \cdot \exp(-(t-t_{last}(\boldsymbol{x}))/\eta) $
and the normalized TS without polarity  
$T_{np}(\boldsymbol{x},t)=255.0 \cdot (T^{'}-\min(T^{'}))/(\max(T^{'})-\min(T^{'}))$
, which are generated from the SAE for point \& line feature tracking and loop closure detection, respectively.
More detailed discussions of these two kinds of event representations can be seen in APPENDIX \ref{Ablation experiment} and Ref.\cite{GWPHKU:MyEVIO}
$ \left( ii \right)$ The EIO Back-end tightly fuses the point landmarks, line landmarks, and the IMU pre-integration to estimate the 6-DoF state, while the loop closure is used to eliminate the accumulated drifts.
Finally, to achieve low latency, we also directly forward propagate (loosely-coupled) the latest estimation with the IMU measurements to achieve IMU-rate state outputs which can be up to 1000 Hz. This can ensure the requirement of closed-loop autonomous quadrotor flight.

For the keyframe in the sliding window, it is selected by two criteria and only based on the event-corner features:
$ \left( i \right)$ When the average parallax of the tracked event-corner features between two consecutive timestamps exceeds a threshold (10 is set in our experiment).
$ \left( ii \right)$ When the number of successfully tracked event-corner features from the last timestamp falls below a certain threshold (20 is set in our experiment).

As for the initialization procedure of our framework, which is adopted from \cite{GWPHKU:VINS-MONO} and \cite{GWPHKU:VINS-MONO-initialization}, our pipeline commences with a vision-only structure from motion (SfM) to establish the up-to-scale structure of camera pose and event-corner feature positions.
By loosely aligning the SfM with the pre-integrated IMU measurements, it can bootstrap the system from unknown initial states rather than using marker \cite{chamorro2022event} or assuming the local scene is planar to the sensor \cite{GWPHKU:EVO}.
It is worth mentioning that if the image is available in the framework, we only employ the point-based image visual measurement for SfM initialization to ensure reliable visual-inertial alignment and up-to-scale camera poses.

Regarding loop closure, extra event corners are detected in the EIO Front-end, subsequently described by the BRIEF descriptor, and fed to the Back-end.
These additional event-corner features are used to achieve a better recall rate on loop detection. 
Thanks to our designed normalized TS without polarity, which is triggered in scenes with strong edges, it can eliminate accumulated drifts and ensure global consistency.
The correspondences are found through the BRIEF descriptor matching by calculating Hamming distance.
When the number of corresponding event descriptors is greater than a certain threshold (16-25 in our experiments), the loop closure is detected.
After detecting the loop, the connection residual of the previous keyframe and the current keyframe are integrated into the nonlinear optimization as a re-localization residual.

We further extend our PL-EIO framework to include point-based image features to provide a more robust state estimation (PL-EVIO). 
Fig.\ref{Point_feature} shows the complementarity of the image and event information. For the bad lighting area, the event can provide reliable event-corner features, while the image can provide rich point-based features in other areas. 
This enables the uniform distribution of the point-based event and image features in the scene.
While the line-based event feature can provide more constraints (shown in Fig.\ref{Line_feature}) even when the successfully tracked point-based event and image features are less in the scene.
Our framework can provide a more robust and accurate state estimation.
More details of event-based point and line feature detection and tracking in our framework can be seen in the APPENDIX \ref{Ablation experiment}.

\subsection{Motion Compensation for the Event Stream using IMU Measuremnet} 
\label{Motion Compensation for the Event Stream using IMU Measuremnet} 
Events can be triggered either by moving objects or by the ego-motion of the camera.
Similar to Ref.\cite{ETH-Science-Robotics}, we only rely on the IMU for motion compensation, which guarantees efficiency and speed.
For the new event stream coming, we use the angular velocity and linear acceleration from the IMU averaged over the time window where the events are grouped in the same event stream, to estimate the ego-rotation and ego-translation of each event. 
Using this ego-motion to warp the events into the timestamp of the first event in the same event stream.
The motion (considering both rotation $\boldsymbol{R}$ and translation $\boldsymbol{T}$) of each event can be calculated through:
\begin{equation}
\boldsymbol{\Delta}(\delta t)=\begin{bmatrix} \boldsymbol{R}(\boldsymbol{\omega}_{IMU}\delta t) & \boldsymbol{T} \\ 0 & 1 \end{bmatrix}
=\begin{bmatrix} \boldsymbol{R}(\boldsymbol{\omega}_{IMU}\delta t) & \frac{1}{2} \boldsymbol{\alpha}_{IMU} \delta t^{2} \\ 0 & 1 \end{bmatrix}
\end{equation}
where
$\boldsymbol{\omega}_{IMU}$ and $\boldsymbol{\alpha}_{IMU}$ are the angular velocity and linear acceleration measurements from the IMU in the current event stream timestamp.
While $\boldsymbol{R}(\boldsymbol{\omega}_{IMU}\delta t)$ is the rotation matrix generated from the angular velocity $\boldsymbol{\omega}_{IMU}$ and the time difference $\delta t$.
Each event $e_{i}=\left\{et_{i},ex_{i},ey_{i},ep_{i} \right\}$ of the event stream is then warped by $\boldsymbol{\Delta}(\delta t)=\boldsymbol{\Delta}(et_{i}-t_{first\_event})$, where $t_{first\_event}$ is the timestamp of the first event of the current event stream and 
$et_{i}$ is the timestamp of event $e_{i}$. 

\subsection{Event-corner Feature Detection and Tracking} 
\label{Event-corner Feature Detection and Tracking} 
\begin{figure}[htb]  
        \vspace{-1.0em}
        \setlength{\abovecaptionskip}{-0.1em}
    	\subfigtopskip=0pt 
    	\subfigbottomskip=0pt 
    	\subfigcapskip=-6pt 
        \centering
        \captionsetup{justification=justified}
        \subfigure[ ]{
                \begin{minipage}[t]{0.47\columnwidth}
                \centering
                \includegraphics[width=1.0\columnwidth]{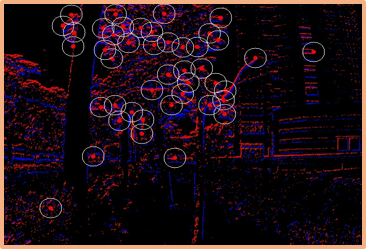}
                \label{event_corner_in_event}
                \end{minipage}%
        }%
        \subfigure[ ]{
                \begin{minipage}[t]{0.47\columnwidth}
                \centering
                \includegraphics[width=1.0\columnwidth]{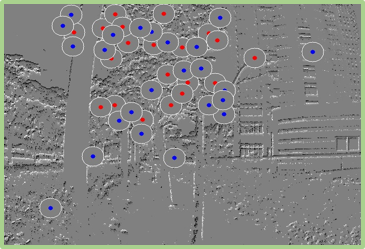}
                \label{event_corner_in_ts}
                \end{minipage}%
        }
        \subfigure[ ]{
                \begin{minipage}[t]{0.94\columnwidth}
                \centering
                \includegraphics[width=1.0\columnwidth]{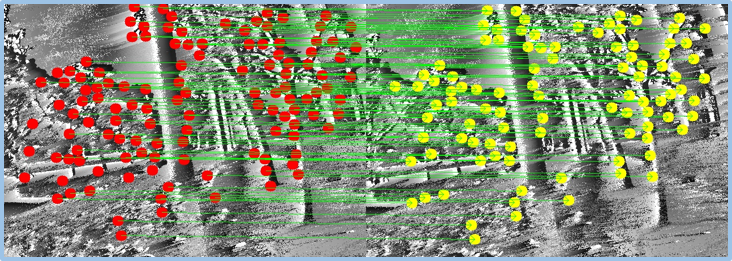}
                \label{event_corner_tracking_in_ts}
                \end{minipage}%
        }
        \caption{The event-corner feature detection and tracking: (a) Detecting features from raw event streams; (b) Using the TS with polarity as the mask for uniform distribution of the event-corner features; (c)Tracking feature in the TS with polarity.}
        \label{event_corner_detection_tracking}
        \vspace{-1.0em}
\end{figure}

The SAE would be updated through the motion-compensated event stream, while the existing event-corner features are tracked by the LK optical flow on the TS with polarity $T_{p}(\boldsymbol{x},t)$ which is generated from the updated SAE (shown in Fig.\ref{event_corner_tracking_in_ts}). 
Different from our previous EIO\cite{GWPHKU:MyEVIO}, in this work, we use a two-way tracking strategy to track event-corner features between two consecutive timestamps.
For any event-corner feature $F_{e}$ on last timestamp $T_{p}(\boldsymbol{x},t)$ is tracked to $F_{e}^{'}$ on current timestamp $T_{p}(\boldsymbol{x},t)$, 
we would reverse the tracking process by tracking $F_{e}^{'}$ on current timestamp $T_{np}(\boldsymbol{x},t)$ back to $F_{e}^{''}$ on last timestamp $T_{p}(\boldsymbol{x},t)$.
If the distance between $F_{e}^{''}$ and $F_{e}^{'}$ is smaller than a threshold (1.0 pixel in our experiment), 
this event-corner feature would be viewed as being successfully tracked.
The event-corner features that are not successfully tracked in the current timestamp would be discarded immediately. 

Whenever the number of the tracked features falls below a certain threshold (150-250 in our experiment), new event-corner features would be detected from the latest motion-compensated event stream (shown in Fig.\ref{event_corner_in_event}) for future feature tracking.
Modified from the publicly available implementation of the Arc* algorithm \cite{GWPHKU:ARC*} for event-based corner detection, we extract the event corners on the individual event by leveraging the SAE rather than adopting the conventional corner detection algorithms in frame-like accumulation (like Ref. \cite{GWPHKU:ETH-EVIO} \cite{GWPHKU:Ultimate-SLAM}).
The newly detected event-corner features would be further selected by setting the TS with polarity as the mask (shown in Fig.\ref{event_corner_in_ts}). 
To enforce the uniform distribution, a minimum distance (10-20 pixels for different resolution event cameras) is set between two neighboring event-corner features. 
Meanwhile, we maintain the event corners, where the pixel value of the TS with polarity is not equal to 128.0, to emphasize the detected event-corner features located in the strong edges rather than the 
numerous noisy features in low texture areas.

Furthermore, all the event-corner features in the front-end are undistorted based on the camera distortion model and projected to a normalized camera coordinate system. 
To remove outliers, we also use the Random Sample Consensus (RANSAC) for outliers filtering. 
Finally, we recover the inverse depth of the event-corner features that are successfully tracked between two consecutive timestamps through triangulation. 
The point-based landmark whose 3D position has been successfully calculated would be fed to the sliding window for nonlinear optimization.

\subsection{Line-based Feature Detection and Matching on Event Stream} 
\label{Line-based Feature Detection and Matching on Event Stream} 
Utilizing the line-based features to improve the performance of point-based VIO is effective as line features can provide additional constraints and structure information in the scene, especially for the human-made environment.
Therefore, for incoming new event streams, after using motion compensation, the streams are mapped into the Opencv-Mat format (event mat).
Given that events are typically triggered in scenes with strong edges, generating line features using the event mat can prevent the generation of invalid line features and enhance efficiency. 
To efficiently extract line-based features and descriptors from the raw event streams, we have modified the LSD algorithm \cite{GWPHKU:LSD} in Opencv.
Utilizing the Sobel filter, we compute the orientation of each event in the event mat and group events with similar angles into a line support region. 
Additionally, we have studied the hidden parameter tuning and length rejection strategy of the LSD algorithm, drawing inspiration from\cite{GWPHKU:PL-VINS} to filter out short line features using a length rejection strategy:
\begin{equation}
L_{\min}=\eta \cdot \min(W,H)
\end{equation}
where $\min(W,H)$ denotes the smaller value between the width and the height of the event camera. $\eta$ is the ratio factor (0.125 is set in our experiments).
After that, we adopt the Line Band Descriptor (LBD)\cite{GWPHKU:LBD} to describe and match line features, respectively. 
In particular, to ensure good tracking performance and be consistent with point-based event-corner features, we also use the TS with polarity for LBD generation and line-based feature matching. 
We further execute the line features refinement schemes to identify line features as good matches for successful line tracking:
\begin{itemize}
\item The Hamming distance between matching line features is less than 30; 
\item The square error of the endpoint between the matching line features is less than 20 * 20 $pixel^{2}$;
\item The angular between matching line features is less than 0.1 rad;
\end{itemize}

The successfully tracked line-based event features would be further refined by undistorting the endpoints of the lines and projected onto a unit sphere after passing outlier rejection. 
The outlier rejection is performed using RANSAC with a fundamental matrix model.
Then, we obtain the line-based landmark by triangulating the correspondences of two line features. 
The line-based landmark whose 3D position has been successfully calculated would be fed to the sliding windows for the nonlinear optimization.

\subsection{Sliding Windows Graph-based Optimization based on the Point-Line Features} 
\label{Sliding Windows Graph-based Optimization based on the Point-Line Features} 

\subsubsection{Formulation of the Nonlinear Optimization} 
\label{Formulation of the Nonlinear Optimization}
The full state vector in the sliding windows is defined as: 
\begin{equation}
\boldsymbol{\chi}=[\boldsymbol{\chi}_{b}, \boldsymbol{\lambda}_{e}, \boldsymbol{\phi}_{l}, \boldsymbol{\lambda}_{c}, \boldsymbol{T}^{b}_{c}, \boldsymbol{T}^{b}_{e}]
\label{state}
\end{equation}
where
$\boldsymbol{\lambda}_{e}=[\lambda_{0}, ..., \lambda_{m_{event}^{th}}]$, and $\boldsymbol{\lambda}_{c}=[\lambda_{0}, ..., \lambda_{m_{image}^{th}}]$ is the inverse depth of the $m_{event}^{th}$ event-corner features and $m_{image}^{th}$ point-based image features, respectively,
while
$\boldsymbol{\phi}_{l}=[\phi_{0}, ..., \phi_{m_{line}^{th}}]$,  
$\phi_{m_{line}^{th}}=[\boldsymbol{\theta}^T,o]$ is the four-parameter orthonormal representation ( as shown in Eq.\eqref{orthonormal_representation_1} and Eq.\eqref{orthonormal_representation_2}) of the $m_{line}^{th}$ line-based event features, in the sliding windows.
$\boldsymbol{T}^{b}_{c}=[\boldsymbol{R}^{b}_{c},\boldsymbol{t}^{b}_{c}]$ or $\boldsymbol{T}^{b}_{e}=[\boldsymbol{R}^{b}_{e},\boldsymbol{t}^{b}_{e}]$ is the extrinsic transformation from camera frame (the image $c$ or event $e$) to the body (IMU) frame $b$
($\boldsymbol{T}^{b}_{c}=\boldsymbol{T}^{b}_{e}$ when using the DAVIS which can simultaneously output the image and event data);
$ \boldsymbol{\chi}_{b}=[\boldsymbol{X}_{1} ,..., \boldsymbol{X}_{K}] $ is the optimization vector in the sliding windows, which comprises the state of the IMU, with $K$ ($K=10$ in our experiments), the total number of keyframes in the sliding windows. 
The system state $\boldsymbol{X}_{k}$ at $k^{th}$ keyframe is given by the position $ \boldsymbol{p}^{w}_{b_{k}} $, orientation quaternion $ \boldsymbol{q}^{w}_{b_{k}} $, and the velocity $ \boldsymbol{v}^{w}_{b_{k}} $ of the IMU in the world frame, and the accelerometer bias $\boldsymbol{b}_{a_{k}}$ and gyroscope bias $\boldsymbol{b}_{g_{k}}$ as follows:
\begin{equation}
\boldsymbol{X}_{k}=[\boldsymbol{p}^{w}_{b_{k}}, \boldsymbol{q}^{w}_{b_{k}}, \boldsymbol{v}^{w}_{b_{k}}, \boldsymbol{b}_{a_{k}}, \boldsymbol{b}_{g_{k}}]
\label{state_body}
\end{equation}
Joint nonlinear optimization is solved for the maximum a posteriori estimation of $\boldsymbol{\chi} $, while the cost function can be written as:
\begin{equation}
\begin{aligned}
       \boldsymbol{J}(\boldsymbol{\chi})=
       &
       \sum_{k=0}^{K-1}\sum_{l\in \zeta}||\boldsymbol{e}^{k,l}_{event}||^{2}_{W^{k}_{event}}+
       \sum_{k=0}^{K-1}\sum_{l\in \ell}||\boldsymbol{e}^{k,l}_{line}||^{2}_{W^{k}_{line}} \\
       &
       +\sum_{k=0}^{K-1}||\boldsymbol{e}^{k}_{imu}||^{2}_{W^{k}_{imu}}+
       \sum_{k=0}^{K-1}\sum_{l\in \xi}||\boldsymbol{e}^{k,l}_{image}||^{2}_{W^{k}_{image}}\\
       &
       +||\boldsymbol{e}_{m}||^{2}_{W_{m}}+||\boldsymbol{e}_{r}||^{2}_{W_{r}}
\label{Joint_nonlinear_optimization}
\end{aligned}
\end{equation}

Eq.\eqref{Joint_nonlinear_optimization} contains 
the point-based event residual $\boldsymbol{e}^{k,l}_{event}$ with weight $W^{k}_{event}$;
the line-based event residual $\boldsymbol{e}^{k,l}_{line}$ with weight $W^{k}_{line}$;
the IMU pre-integration residuals $\boldsymbol{e}^{k}_{imu}$ with weight $W^{k}_{imu}$;
the point-based image residual $\boldsymbol{e}^{k,l}_{image}$ with weight $W^{k}_{image}$;
the marginalization residuals $\boldsymbol{e}_{m}$ with weight $W_{m}$; 
the re-localization residuals $\boldsymbol{e}_{r}$ with weight $W_{r}$;
while $\zeta$, $\ell$, and $\xi$ are the set of event-corner features, line-based event features, and point-based event features, respectively, which have been successfully tracked or matched at least twice in the current sliding window.

\subsubsection{Point-based Event Visual Measurement Residual} 
\label{Point-based Event Visual Measurement Residual}
The $\boldsymbol{e}^{k,l}_{event}$ in Eq.\eqref{Joint_nonlinear_optimization} is the event-corner measurement residual from the re-projection function. 
Considering the $l^{th}$ event-corner feature that is first observed in the $i^{th}$ keyframe, the residual for its observation in the $k^{th}$ keyframe is defined as:
\begin{equation}
\vspace{-0.2cm} 
\begin{split}
        \boldsymbol{e}^{k,l}_{event}=\left[
                \begin{array}{c}
                u^{l}_{k}\\
                v^{l}_{k}\\
                \end{array} 
                \right]-\pi_{e}\cdot(\boldsymbol{T}^{b}_{e})^{-1}\cdot &\boldsymbol{T}^{b_{k}}_{w}\cdot \boldsymbol{T}^{w}_{b_{i}}\cdot \boldsymbol{T}^{b}_{e}\\
                &\cdot \pi^{-1}_{e}(\frac{1}{\lambda_{e}},\left[
                        \begin{array}{c}
                        u^{l}_{i}\\
                        v^{l}_{i}\\
                        \end{array} 
                        \right])
 \label{re-projection-point-event}
\end{split}
\end{equation}
where, 
$\left[ u^{l}_{i}, v^{l}_{i} \right]^{\mathrm{T}} $ 
is the first observation of the $l^{th}$ event-corner feature in the $i^{th}$ keyframe. 
$\left[ u^{l}_{k}, v^{l}_{k} \right]^{\mathrm{T}} $ 
is the observation of the same event-corner feature in the $k^{th}$ keyframe, $\pi_{e}$ and $\pi^{-1}_{e}$ are the projection and back-projection function of the event camera, respectively, which include the intrinsic parameters for the transform between the 2D pixel coordinates and normalized event camera coordinate.
$\boldsymbol{T}^{w}_{b_{i}}$ indicates the movement of the body frame related to the world frame in timestamp $i$, $\boldsymbol{T}^{b_{k}}_{w}$ is the transpose of the pose of the body in the world frame in the $k^{th}$ keyframe.

\subsubsection{Line-based Event Visual Measurement Residual} 
\label{Line-based Event Visual Measurement Residual}
The $\boldsymbol{e}^{k,l}_{line}$ in Eq.\eqref{Joint_nonlinear_optimization} is the line-based event measurement residual which is generated from line re-projection model.
The line re-projection residual is modeled as the distance from the endpoints of the line to the projected line in the normalized image plane.
The $l^{th}$ line-based landmarks in the world frame can be defined using the Plücker Coordinate: $\boldsymbol{L}^{l}_{w}=
\left[
        \begin{array}{c}
        \boldsymbol{n}^{l}_{w}, 
        \boldsymbol{d}^{l}_{w}\\
        \end{array} 
        \right]^{\mathrm{T}}$
, $\boldsymbol{n}^{l}_{w}$ denotes the normal vector of the plane determined by $\boldsymbol{L}^{l}_{w}$ and the origin of the world frame, 
while $\boldsymbol{d}^{l}_{w}$ denotes the direction vector determined by the two endpoints of $\boldsymbol{L}^{l}_{w}$.
Given the transformation matrix $\boldsymbol{T}^{b_{k}}_{w}=\left[ \boldsymbol{R}^{b_{k}}_{w}, \boldsymbol{t}^{b_{k}}_{w}\right]$ indicates the movement of the body frame related to the world frame in timestamp $k$, we can obtain the transformation from the world frame to the event frame in timestamp $k$ through $\boldsymbol{T}^{e_{k}}_{w}=\boldsymbol{T}^{e}_{b} \cdot \boldsymbol{T}^{b_{k}}_{w}$, where $\boldsymbol{T}^{e}_{b}=[\boldsymbol{R}^{e}_{b},\boldsymbol{t}^{e}_{b}]$ is the extrinsic transformation from the body (IMU) frame $b$ to the event camera frame $e$.
Then, we can transform the $l^{th}$ line-based event feature $\boldsymbol{L}^{l}_{w}$ in $k^{th}$ keyframe from world frame to event camera frame by\cite{zhang2015building}\cite{bartoli20043d}:
\begin{equation}
\begin{aligned}
\boldsymbol{L}^{l}_{e_{k}}=
\left[
        \begin{array}{c}
        \boldsymbol{n}^{l}_{e_{k}}\\
        \boldsymbol{d}^{l}_{e_{k}}\\
        \end{array} 
        \right]
&  =
\left[
    \begin{matrix}
    \boldsymbol{R}^{e_{k}}_{w} & [\boldsymbol{t}^{e_{k}}_{w}] \times \boldsymbol{R}^{e_{k}}_{w} \\
    0     & \boldsymbol{R}^{e_{k}}_{w}\\
     \end{matrix}
\right] 
\left[
        \begin{array}{c}
        \boldsymbol{n}^{l}_{w}\\
        \boldsymbol{d}^{l}_{w}\\
        \end{array} 
        \right] \\
\label{line-1}
\end{aligned}
\end{equation}
where $\boldsymbol{R}^{e_{k}}_{w}=\boldsymbol{R}^{e}_{b} \cdot \boldsymbol{R}^{b_{k}}_{w},  \boldsymbol{t}^{e_{k}}_{w}=\boldsymbol{R}^{e}_{b} \cdot \boldsymbol{t}^{b_{k}}_{w}+\boldsymbol{t}^{e}_{b}.$

The transformation for the Plücker Coordinates of the $l^{th}$ line-based event feature from $i^{th}$ keyframe to $k^{th}$ keyframe in the body frame can be represented as follows:
\begin{equation}
\begin{aligned}
\boldsymbol{L}^{l}_{b_{k}}=
\left[
        \begin{array}{c}
        \boldsymbol{n}^{l}_{b_{k}}\\
        \boldsymbol{d}^{l}_{b_{k}}\\
        \end{array} 
        \right]
&  =
\left[
    \begin{matrix}
    \boldsymbol{R}^{b_{k}}_{b_{i}} & [\boldsymbol{t}^{b_{k}}_{b_{i}}] \times \boldsymbol{R}^{b_{k}}_{b_{i}} \\
    0     & \boldsymbol{R}^{b_{k}}_{b_{i}}\\
     \end{matrix}
\right] 
\left[
        \begin{array}{c}
        \boldsymbol{n}^{l}_{b_{i}}\\
        \boldsymbol{d}^{l}_{b_{i}}\\
        \end{array} 
        \right] \\
\label{line-2}
\end{aligned}
\end{equation}
where $\boldsymbol{T}^{b_{k}}_{b_{i}}=\left[ \boldsymbol{R}^{b_{k}}_{b_{i}}, \boldsymbol{t}^{b_{k}}_{b_{i}}\right]$ indicates the movement of the body frame related to the world frame in $i^{th}$ keyframe to $k^{th}$ keyframe.

The Plücker Coordinates $\boldsymbol{L}^{l}_{w}$ can be represented using a four-parameter orthonormal representation, known for its superior convergence performance\cite{zhang2015building}.
As a result, we transfer the line-based landmark to the four-parameter orthonormal representation for the optimization process.
The orthonormal representation $(\boldsymbol{U},\boldsymbol{W}) \in (SO(3), SO(2)) $ of the Plücker Coordinates
$\boldsymbol{L}^{l}_{w}$
can be computed using the QR decomposition\cite{zhang2015building} \cite{GWPHKU:PL-VINS}:
\begin{equation}
[\boldsymbol{n}^{l}_{w}|\boldsymbol{d}^{l}_{w}]=\boldsymbol{U} 
\left[
    \begin{matrix}
    w_{1} & 0\\
    0     & w_{2}\\
    0     & 0 \\
     \end{matrix}
\right] , set:
\boldsymbol{W} =
\left[
    \begin{matrix}
    w_{1} & -w_{2}\\
    w_{2}    & w_{1}\\
     \end{matrix}
\right]
\end{equation}
where $\boldsymbol{U}$ and $\boldsymbol{W}$ denote a three and a two dimensional rotation matrix, respectively.
Let $\boldsymbol{R}(\boldsymbol{\theta})=\boldsymbol{U}$ and $\boldsymbol{R}(o)=\boldsymbol{W}$ be the corresponding rotation transformations, where $\boldsymbol{U}=\left[ u_{1}, u_{2}, u_{3}\right]$.
With this notation, we can now express the relationship as follows:
\begin{equation}
\boldsymbol{R}(\boldsymbol{\theta})=\boldsymbol{U} = 
\left[
    \frac{\boldsymbol{n}^{l}_{w}}{||\boldsymbol{n}^{l}_{w}||},   \frac{\boldsymbol{d}^{l}_{w}}{||\boldsymbol{d}^{l}_{w}||},  \frac{\boldsymbol{n}^{l}_{w}\times \boldsymbol{d}^{l}_{w}}{||\boldsymbol{n}^{l}_{w} \times \boldsymbol{d}^{l}_{w}||}
\right]
\label{orthonormal_representation_1}
\end{equation}

\begin{equation}
\begin{aligned}
\boldsymbol{R}(o)&=\boldsymbol{W}
=
\left[
    \begin{matrix}
    cos(o) & -sin(o)\\
    sin(o)    & cos(o)\\
     \end{matrix}
\right] \\
& =
\frac{1}{\sqrt{||\boldsymbol{n}^{l}_{w}||^{2}+||\boldsymbol{d}^{l}_{w}||^{2}}}
\left[
    \begin{matrix}
   ||\boldsymbol{n}^{l}_{w}|| & -||\boldsymbol{d}^{l}_{w}||\\
    ||\boldsymbol{d}^{l}_{w}||    & ||\boldsymbol{n}^{l}_{w}||\\
     \end{matrix}
\right]
\\
\end{aligned}
\label{orthonormal_representation_2}
\end{equation}
Up to this point, we have established the connection between the four-parameter orthonormal representation $\phi_{m_{line}^{th}}=[\boldsymbol{\theta}^\mathrm{T},o]$ of Eq.\eqref{state} and the Plücker Coordinates $\boldsymbol{L}^{l}_{w}$.

The Plücker Coordinates $\boldsymbol{L}^{l}_{e_{k}}$ in the event camera frame can be obtained from $\boldsymbol{L}^{l}_{w}$ through Eq.\eqref{line-1}, and then can be projected to the line $\boldsymbol{l}^{l}_{e_{k}}$ in the event imaging plane by
\begin{equation}
\boldsymbol{l}^{l}_{e_{k}}=
\left[
\begin{array}{c}
l_{1},
l_{2},
l_{3}\\
\end{array} 
\right]^\mathrm{T}= 
\pi_{e} \boldsymbol{n}^{l}_{e_{k}}
\end{equation}
where $\pi_{e}$ is the projection function of the event camera, and the $\boldsymbol{n}^{l}_{e_{k}}$ can be obtained from Eq.\eqref{line-1}.
The line re-projection error in Eq.\eqref{Joint_nonlinear_optimization} can be defined as:
\begin{equation}
\boldsymbol{e}^{k,l}_{line}=\begin{bmatrix} d(\boldsymbol{S}_{\boldsymbol{l}^{l}_{e_{k}}},\boldsymbol{l}^{l}_{e_{k}}) \\ d(\boldsymbol{E}_{\boldsymbol{l}^{l}_{e_{k}}},\boldsymbol{l}^{l}_{e_{k}})  \end{bmatrix}
\end{equation}
where $\boldsymbol{S}_{\boldsymbol{l}^{l}_{e_{k}}}$ and $\boldsymbol{E}_{\boldsymbol{l}^{l}_{e_{k}}}$ are the homogeneous coordinates of the endpoints of the line feature $\boldsymbol{l}^{l}_{e_{k}}$ in the image plane,
and $d(m,\boldsymbol{l}^{l}_{e_{k}})$ denotes the point-to-line distance function from the endpoints to the projection line $\boldsymbol{l}^{l}_{e_{k}}$:
\begin{equation}
\begin{aligned}
&
d(m,\boldsymbol{l}^{l}_{e_{k}})=
\frac{m\boldsymbol{l}^{l}_{e_{k}}}{\sqrt{l_{1}^{2}+l_{2}^{2}}} \\
&
\boldsymbol{S}_{\boldsymbol{l}^{l}_{e_{k}}}=(u^{l}_{k,S}, v^{l}_{k,S}, 1),  \boldsymbol{E}_{\boldsymbol{l}^{l}_{e_{k}}}=(u^{l}_{k,E}, v^{l}_{k,E},1) \\
\end{aligned}
\end{equation}

\subsubsection{Point-based Image Visual Measurement Residual}
\label{Point-based Image Visual Measurement Residual}

The $\boldsymbol{e}^{k,l}_{image}$ in Eq.\eqref{Joint_nonlinear_optimization} is the point-based image measurement residual from the re-projection function. 
Similar to the event-corner measurement, the $l^{th}$ point-based image feature that is first observed in the $i^{th}$ keyframe, the residual for its observation in the $k^{th}$ keyframe is defined as:
\begin{equation}
\begin{split}
        \boldsymbol{e}^{k,l}_{image}=\left[
                \begin{array}{c}
                u^{l}_{k}\\
                v^{l}_{k}\\
                \end{array} 
                \right]-\pi_{c}\cdot(\boldsymbol{T}^{b}_{c})^{-1}\cdot &\boldsymbol{T}^{b_{k}}_{w}\cdot \boldsymbol{T}^{w}_{b_{i}}\cdot \boldsymbol{T}^{b}_{c}\\
                &\cdot \pi^{-1}_{c}(\frac{1}{\lambda_{c}},\left[
                        \begin{array}{c}
                        u^{l}_{i}\\
                        v^{l}_{i}\\
                        \end{array} 
                        \right])
 \label{re-projection-point-image}
\end{split}
\end{equation}
where,   
$\left[ u^{l}_{i}, v^{l}_{i} \right]^\mathrm{T}$ 
is the first observation of the $l^{th}$ point-based image feature in the $i^{th}$ keyframe. 
$\left[ u^{l}_{k}, v^{l}_{k} \right]^\mathrm{T}$ 
is the observation of the same point-based image feature in the $k^{th}$ keyframe. $\pi_{c}$ and $\pi^{-1}_{c}$ are the projection and back-projection function of the standard camera, respectively, which include the intrinsic parameters for the transform between the 2D pixel coordinates and normalized camera coordinate.

\subsubsection{IMU Measurement Residual} 
\label{IMU Measurement Residual}

The $\boldsymbol{e}^{k}_{imu}$ in Eq.\eqref{Joint_nonlinear_optimization} is the IMU residual from the IMU pre-integration. The raw measurement of angular velocity $\boldsymbol{\omega}_{k}$ and acceleration $\boldsymbol{a}_{k}$ from IMU at time $t_{k}$ are:
\begin{equation}
\begin{split}
\label{measurement_of_angular_velocity}
&\hat{\boldsymbol{a}}_{k} = \boldsymbol{a}_{k} - \boldsymbol{R}^{b_{k}}_{\boldsymbol{\omega}}\boldsymbol{g}^{\omega} + \boldsymbol{b}_{a_{k}}+\boldsymbol{n}_{a} \\
&\hat{\boldsymbol{\omega}}_{k} = \boldsymbol{\omega}_{k}+\boldsymbol{b}_{\omega_{k}}+\boldsymbol{n}_{\omega}
\end{split}
\end{equation}
where $\boldsymbol{n}_{a}$, $\boldsymbol{n}_{\omega}$ are modeled as additive Gaussian noise. $\boldsymbol{b}_{a_{k}}$, $\boldsymbol{b}_{\omega_{k}}$ are modeled as random walks. The Notation $\hat{(\cdot)}$ is used to represent noisy measurements.
Given the time interval $[t_{k}, t_{k+1}]$ corresponding to keyframe $b_{k}$ and $b_{k+1}$. $\boldsymbol{p}^{\omega}_{b_{k+1}}$, $\boldsymbol{v}^{\omega}_{b_{k+1}}$, $\boldsymbol{q}^{\omega}_{b_{k+1}}$ can be propagated in such time interval by using gyroscope and accelerometer measurements in the world frames as follows:
\begin{equation}
\begin{split}
\label{IMU_motion_model}
&\boldsymbol{p}^{\omega}_{b_{k+1}} = \boldsymbol{p}^{\omega}_{b_{k}}+\boldsymbol{v}^{\omega}_{b_{k}}\Delta t + \iint^{t_{k+1}}_{t_{k}}(\boldsymbol{R}^{\omega}_{b_{k}}\boldsymbol{a}_{k}) \delta t^{2}\\
&\boldsymbol{v}^{\omega}_{b_{k+1}} = \boldsymbol{v}^{\omega}_{b_{k}} + \int^{t_{k+1}}_{t_{k}} (\boldsymbol{R}^{\omega}_{b_{k}}\boldsymbol{a}_{k})\delta t \\
&\boldsymbol{q}^{\omega}_{b_{k+1}} = \int^{t_{k+1}}_{t_{k}} \boldsymbol{q}^{\omega}_{b_{k}} \otimes \left[
                        \begin{array}{c}
                        0 \\
                        \frac{1}{2}\boldsymbol{\omega}_{k}\\
                        \end{array}
                        \right] \delta t
\end{split}
\end{equation}
Based on Eq.\eqref{measurement_of_angular_velocity}, Eq.\eqref{IMU_motion_model} can be rewritten as follows:
\begin{equation}
\begin{split}
\label{IMU_motion_model_with_measurements}
\boldsymbol{p}^{\omega}_{b_{k+1}} - \boldsymbol{p}^{\omega}_{b_{k}} &- \boldsymbol{v}^{\omega}_{b_{k}}\Delta t - \frac{1}{2}\boldsymbol{g}^{\omega}\Delta t^{2}\\ 
&= \iint^{t_{k+1}}_{t_{k}}(\boldsymbol{R}^{\omega}_{b_{k}}(\hat{\boldsymbol{a}}_{k}
-\boldsymbol{b}_{a_{k}}-\boldsymbol{n}_{a})) \delta t^{2}\\
\boldsymbol{v}^{\omega}_{b_{k+1}} - \boldsymbol{v}^{\omega}_{b_{k}}& - \boldsymbol{g}^{\omega}\Delta t\\
&= \int^{t_{k+1}}_{t_{k}} (\boldsymbol{R}^{\omega}_{b_{k}} \hat{\boldsymbol{a}}_{k}\delta t - \boldsymbol{R}^{\omega}_{b_{k}}\boldsymbol{b}_{a_{k}}\delta t-\boldsymbol{R}^{\omega}_{b_{k}} \boldsymbol{n}_{a} \delta t ) \\
\boldsymbol{q}^{\omega}_{b_{k+1}} &= \int^{t_{k+1}}_{t_{k}} \boldsymbol{q}^{\omega}_{b_{k}} \otimes \left[
                        \begin{array}{c}
                        0 \\
                        \frac{1}{2}(\hat{\boldsymbol{\omega}}_{k}-\boldsymbol{b}_{\omega_{k}}-\boldsymbol{n}_{\omega})\\
                        \end{array}
                        \right] \delta t
\end{split}
\end{equation}
In order to ensure the pre-integration term is only related to the inertial measurements and biases in $[t_{k}, t_{k+1}]$, $\boldsymbol{R}^{b_{k}}_{\omega}$ is multiplied on both sides of Eq.\eqref{IMU_motion_model_with_measurements}, and we define the pre-integration term $\boldsymbol{\alpha}^{b_{k}}_{b_{k+1}}$, $\boldsymbol{\beta}^{b_{k}}_{b_{k+1}}$, $\boldsymbol{\gamma}^{b_{k}}_{b_{k+1}}$ as follows:
\begin{equation}
\begin{split}
\label{alpha_beta_gamma}
\boldsymbol{\alpha}^{b_{k}}_{b_{k+1}} &= \boldsymbol{R}^{b_{k}}_{\omega} \iint^{t_{k+1}}_{t_{k}}(\boldsymbol{R}^{\omega}_{b_{k}}(\hat{\boldsymbol{a}}_{k} - \boldsymbol{b}_{a_{k}} - \boldsymbol{n}_{a}))\delta t^{2}\\
\boldsymbol{\beta}^{b_{k}}_{b_{k+1}} &= \boldsymbol{R}^{b_{k}}_{\omega} \int^{t_{k+1}}_{t_{k}}(\boldsymbol{R}^{\omega}_{b_{k}} \hat{\boldsymbol{a}_{k}} \delta t - \boldsymbol{R}^{\omega}_{b_{k}} \boldsymbol{b}_{a_{k}} \delta t - \boldsymbol{R}^{\omega}_{b_{k}} \boldsymbol{n}_{a} \delta t) \\
\boldsymbol{\gamma}^{b_{k}}_{b_{k+1}} &= \boldsymbol{q}^{\omega}_{b_{k}} \otimes \boldsymbol{q}^{\omega}_{b_{k+1}}\\
\end{split}
\end{equation}
Discretizing Eq.\eqref{alpha_beta_gamma} by the zero-order discretization method as follows:
\begin{equation}
\begin{split}
\label{discrete_alpha_beta_gamma}
\hat{\boldsymbol{\alpha}}^{b_{k}}_{b_{i+1}} &= \hat{\boldsymbol{\alpha}}^{b_{k}}_{b_{i}} + \hat{\boldsymbol{\beta}}^{b_{k}}_{b_{i}}\delta t + \frac{1}{2} \boldsymbol{R}(\hat{\gamma}^{b_{k}}_{b_{i}})(\hat{\boldsymbol{a}}_{i} - \boldsymbol{b}_{a_{i}}) \delta t^{2}\\
\hat{\boldsymbol{\beta}}^{b_{k}}_{b_{i+1}} &= \hat{\boldsymbol{\beta}}^{b_{k}}_{b_{i}} + \boldsymbol{R}(\hat{\boldsymbol{\gamma}}^{b_{k}}_{b_{i}})(\hat{\boldsymbol{a}}_{i} - \boldsymbol{b}_{a_{i}}) \delta t \\
\hat{\boldsymbol{\gamma}}^{b_{k}}_{b_{i+1}} &= \hat{\boldsymbol{\gamma}}^{b_{k}}_{b_{i}} \otimes \left[
                        \begin{array}{c}
                        1 \\
                        \frac{1}{2}(\hat{\boldsymbol{\omega}}_{i}-\boldsymbol{b}_{\omega_{i}})\\
                        \end{array}
                        \right] \delta t
\end{split}
\end{equation}

\begin{table*}[htbp] 
        \renewcommand\arraystretch{1.2} 
        \begin{center}
        \caption{Accuracy Comparison of Our PL-EIO with Other Image-based or Event-based VIO Works}
        \label{vicon_Comparison_with_pleio}
        \setlength{\tabcolsep}{1.0mm} 
        \resizebox{\columnwidth*2}{!}
        { 
        \begin{threeparttable}
        \begin{tabular}{c|ccccccccc|cccc} 
        \hline  
        \multirow{2}*{Sequence}   &\multicolumn{9}{c|}{DAVIS346 (346*260)} &\multicolumn{4}{c}{DVXplorer (640*480)} \\
\cline{2-14}
~   & \makecell{VINS-MONO \cite{GWPHKU:VINS-MONO} \\VIO} 
    & \makecell{ORB-SLAM3 \cite{ORB-SLAM3} \\VO} 
    & \makecell{PL-VINS \cite{GWPHKU:PL-VINS} \\VIO} 
    &\makecell{Ultimate SLAM \cite{GWPHKU:Ultimate-SLAM} \\EIO} 
    &\makecell{Ultimate SLAM \cite{GWPHKU:Ultimate-SLAM} \\EVIO} 
    &\makecell{Our EIO \cite{GWPHKU:MyEVIO} \\EIO}
    &\makecell{Our PL-EIO \\EIO}
    &\makecell{Our PL-EIO+ \\EIO}
    &\makecell{Our PL-EVIO \\EVIO}
    &\makecell{Ultimate SLAM \cite{GWPHKU:Ultimate-SLAM} \\EIO} 
    &\makecell{Our EIO \cite{GWPHKU:MyEVIO} \\EIO}
    &\makecell{Our PL-EIO \\EIO} 
    &\makecell{Our PL-EIO+ \\EIO} \\  
\hline
vicon\_hdr1            & 0.96           &0.32            &0.67            &1.49            &2.44 & 0.59 & 0.67 &0.57 &0.17          &1.94            & 0.30 & 0.47 & 0.41\\
vicon\_hdr2            & 1.60           &0.75            &0.90            &1.28            &1.11 & 0.74 & 0.45 &0.54 &0.12          &2.38            & 0.37 & 0.22 & 0.21\\
vicon\_hdr3            & 2.28           &0.60            &0.69            &0.66            &0.83 & 0.72 & 0.74 &0.69 &0.19          &0.83            & 0.69 & 0.47 & 0.36\\
vicon\_hdr4            & 1.40           &0.70            &0.66            &1.84            &1.49 & 0.37 & 0.37 &0.32 &0.11          &2.09            & 0.26 & 0.27 & 0.25\\
vicon\_darktolight1    & 0.51           &0.75            &0.84            &1.33            &1.00 & 0.81 & 0.78 &0.66 &0.14          &1.96            & 0.80 & 0.71 & 0.71\\
vicon\_darktolight2    & 0.98           &0.76            &1.50            &1.48            &0.79 & 0.42 & 0.44 &0.51 &0.12          &1.57            & 0.57 & 0.56 & 0.47\\
vicon\_lighttodark1    & 0.55           &0.41            &0.64            &1.79            &0.84 & 0.29 & 0.42 &0.33 &0.13          &2.48            & 0.81 & 0.43 & 0.54\\
vicon\_lighttodark2    & 0.55           &0.58            &0.93            &1.32            &1.49 & 0.79 & 0.73 &0.53 &0.16          &1.37            & 0.75 & 0.67 & 0.60\\
vicon\_dark1           & 0.88           &\textit{failed} &0.53            &1.75            &3.45 & 1.02 & 0.64 &0.35 &0.43          &3.79            & 0.35 & 0.51 & 0.41\\
vicon\_dark2           & 0.52           &0.60            &\textit{failed} &1.10            &0.63 & 0.49 & 0.30 &0.38 &0.47          &2.81            & 0.41 & 0.38 & 0.41\\
vicon\_aggressive\_hdr &\textit{failed} &\textit{failed} &1.94            &\textit{failed} &2.30 & 0.66 & 0.62 &0.50 &1.97          &\textit{failed} & 0.65 & 0.62 & 0.50\\
\hline 
Average                & 1.02           &0.61            &0.93            &1.40            &1.49 & 0.63 & 0.56 &0.49 &\textbf{0.36} &2.12            & 0.54 &0.48  & \textbf{0.45}\\
\hline        
        \end{tabular}
        \begin{tablenotes} 
        \item \textit{Unit:\%/m, 0.45 means the average error would be 0.45m for 100m motion.} 
        \end{tablenotes} 
        \end{threeparttable} 
        }
        \end{center}
        \vspace{-2.0em}
\end{table*}

Eventually, the IMU residual can be derived as follows:
\begin{equation}
\begin{split}
\label{IMU_residual}
\boldsymbol{e}^{k}_{imu} =
\left[
\begin{array}{c}
\boldsymbol{R}^{b_{k}}(\boldsymbol{p}^{\omega}_{b_{k+1}}-\boldsymbol{p}^{\omega}_{b_{k}}-\boldsymbol{v}^{\omega}_{b_{k}}\Delta t - \frac{1}{2}\boldsymbol{g}^{\omega}\Delta t^{2})- \hat{\boldsymbol{\alpha}}^{b_{k}}_{b_{k+1}}  \\
\boldsymbol{R}^{b_{k}}(\boldsymbol{v}^{\omega}_{b_{k+1}}-\boldsymbol{v}^{\omega}_{b_{k}}-\boldsymbol{g}^{\omega}\Delta t)- \hat{\boldsymbol{\beta}}^{b_{k}}_{b_{k+1}}\\
2\left[ (\boldsymbol{q}^{\omega}_{b_{k}})^{-1} \otimes \boldsymbol{q}^{\omega}_{b_{k+1}} \otimes (\hat{\boldsymbol{\gamma}}^{b_{k}}_{b_{k+1}})^{-1} \right]_{xyz} \\
\boldsymbol{b}_{a_{k+1}}-\boldsymbol{b}_{a_{k}}\\
\boldsymbol{b}_{\omega_{k+1}}-\boldsymbol{b}_{\omega_{k}}\\
\end{array}
\right]
\end{split}
\end{equation}

\section{Evaluation} 
\label{Evaluation}

In this section, we evaluate the effectiveness of our framework in various challenging sequences using both quantitative and qualitative methods in subsection \ref{Evaluation in High-Dynamic-Range Scenarios} and \ref{Evaluation in Aggressive Motions}. 
We implemented our method with C++ in Ubuntu 20.04 and ROS Noetic. 
All sequences are evaluated in real-time using a laptop with Intel Core i7-11800H and are recorded in videos (shown on our project website). 
In subsection \ref{Online Quadrotor-flight Evaluation} and \ref{Aggressive Quadrotor-flip Evaluation}, we demonstrate the quadrotor flight using our method for the closed-loop state estimator and aggressive flip. 
Meanwhile, large-scale experiments are carried out to illustrate the long-time practicability in subsection \ref{Outdoor Large-scale Evaluation}. 

\subsection {Evaluation in High-Dynamic-Range Scenarios} 
\label{Evaluation in High-Dynamic-Range Scenarios}

For demonstrating the robustness, accuracy, and real-time capability, we initially evaluate our PL-EIO using different resolution event cameras (DAVIS346 (346*260) and DVXplorer (640*480)) with the ground truth from VICON. 
All sequences\footnote{\url{https://github.com/arclab-hku/Event_based_VO-VIO-SLAM}} are recorded in broad illumination range conditions, or under aggressive motion.
Without loss of generality, we use the raw image from DAVIS346 to run the VINS-MONO \cite{GWPHKU:VINS-MONO}, PL-VINS \cite{GWPHKU:PL-VINS}, and ORB-SLAM3 \cite{ORB-SLAM3}, as image-based comparisons. 
In addition, based on the source code of Ultimate SLAM \cite{GWPHKU:Ultimate-SLAM}, we also test the EVIO and EIO versions of Ultimate SLAM for event-based comparison.
The estimated and ground-truth trajectories are aligned with a 6-DOF transformation (in SE3), using 5 seconds [0-5s] of the resulting trajectory. We compute the mean position error (Euclidean distance in meters) as percentages of the total travel distance of the ground truth, which is calculated by the publicly available tool \cite{GWPHKU:evo_package}. 
As can be seen from the results in Table \ref{vicon_Comparison_with_pleio}, our PL-EIO has better performances compared with the other methods in different resolution event cameras. Especially, for the results of \textit{vicon\_aggressive\_hdr}, our PL-EIO produces reliable and accurate pose estimation even when the image-based VIO and VO fail.  
Besides, compared with our previous EIO\cite{GWPHKU:MyEVIO}, the introduction of the line feature, known as PL-EIO, demonstrates significant performance improvements across different resolution event cameras. 
While the performance of PL-EVIO, which incorporates image measurements, surpasses our EIO \cite{GWPHKU:MyEVIO}.
Our experimental observations indicate that although the image-aid one (PL-EVIO) exhibits notable performance gains in most sequences, it underperforms in low-light environments such as \textit{vicon\_dark1} and \textit{vicon\_dark2}), as compared to PL-EIO. 
This could be attributed to the degradation of point-based image feature tracking in dark environments.

Regarding the proposed motion compensation algorithm, as evident from the results, the motion compensation version (PL-EIO+) does not exhibit significant enhancements across various sequences, particularly in scenarios involving aggressive motion.
This outcome could potentially stem from biases present in the IMU during such aggressive motion.
On the other hand, in this evaluation, the event stream rate is 60Hz for DAVIS346 and 50Hz for DVXplorer. 
Such high frequencies may lead to reduce time differences within the same event stream, this could potentially be another reason for the limited effectiveness of our motion compensation algorithm.
Additionally, we observe that motion compensation for event streams may not be an optimal choice for high-resolution event cameras due to the trade-off between computational burden and performance improvement.

It is worth mentioning that the Ultimate-SLAM is just for reference since we do not deeply fine-tune the parameters for different sequences (being failure-free is difficult). 
Since the illumination would change greatly in our dataset, and it is very difficult for Ultimate-SLAM to choose a certain stationary threshold to integrate the event stream into the edge image. 
We have tried our best to fine-tune the parameters of Ultimate-SLAM in sequence \textit{vicon\_hdr3} to achieve good performance and use the same parameters to evaluate other sequences. 
This also shows that the generalization ability to integrate the event streams into the edge-image for VIO is pretty bad since the number of triggered events depends on many factors, including the resolution of the camera, the texture of the sense, the illumination, etc.

\subsection{Evaluation in Aggressive Motions} 
\label{Evaluation in Aggressive Motions}

In this section, we evaluate our PL-EVIO in UZH-FPV dataset \cite{GWPHKU:FPV}, which is a high-speed, aggressive visual-inertial odometry dataset. 
This dataset includes fast laps around a racetrack with drone racing gates, as well as free-form trajectories around obstacles. 
We compare our PL-EVIO with ORB-SLAM3 (stereo VIO) \cite{ORB-SLAM3}, VINS-Fusion (stereo VIO) \cite{GWPHKU:VINS-Fusion}, VINS-MONO (monocular VIO) \cite{GWPHKU:VINS-MONO}, and Ultimate SLAM (EVIO) \cite{GWPHKU:Ultimate-SLAM}. 
We also computed the mean position error as percentages of the total traveled distance, while the estimated trajectories and ground-truth were aligned in SE3 with all alignments.
As can be seen from the results in Table \ref{FPV_Comparison_with_evio}, our proposed PL-EVIO achieve better performance even compared with the stereo VIO using a higher resolution camera.
This dataset is so challenging that most of the sequences using Ultimate-SLAM and VINS-Fusion failed, while our Pl-EVIO still can provide reliable and satisfying results. 
To achieve optimal performance, deep fine-tuning of parameters is also required for VINS-MONO.

\begin{table}[htbp] 
        \renewcommand\arraystretch{1.2}
        \begin{center}
        \caption{Accuracy Comparison of Our PL-EVIO with Other Image/Event-based VIO in UZH-FPV Dataset \cite{GWPHKU:FPV}}
        \label{FPV_Comparison_with_evio}
        \resizebox{\columnwidth}{!}
        { 
        \begin{threeparttable}
        \setlength{\tabcolsep}{1.0mm}
        \begin{tabular}{c|cc|ccc} 
        \hline
        \multirow{3}*{Sequence}   &\multicolumn{2}{c|}{Snapdragon (640*480)} &\multicolumn{3}{c}{DAVIS346 (346*260)} \\
\cline{2-6}
~   & \makecell{VINS-Fusion \cite{GWPHKU:VINS-Fusion} \\Stereo VIO} 
    & \makecell{ORB-SLAM3 \cite{ORB-SLAM3} \\Stereo VIO} 
    & \makecell{VINS-MONO \cite{GWPHKU:VINS-MONO} \\VIO} 
    &\makecell{Ultimate SLAM \cite{GWPHKU:Ultimate-SLAM} \\EVIO}
    & \makecell{Our PL-EVIO \\EVIO} \\
\hline
Indoor\_forward\_3      &0.84            &0.55            &0.65 &\textit{failed} &0.38 \\
Indoor\_forward\_5      &\textit{failed} &1.19            &1.07 &\textit{failed} &0.90 \\
Indoor\_forward\_6      &1.45            &\textit{failed} &0.25 &\textit{failed} &0.30 \\
Indoor\_forward\_7      &0.61            &0.36            &0.37 &\textit{failed} &0.55 \\
Indoor\_forward\_9      &2.87            &0.77            &0.51 &\textit{failed} &0.44 \\
Indoor\_forward\_10     &4.48            &1.02            &0.92 &\textit{failed} &1.06 \\
Indoor\_45\_degree\_2   &\textit{failed} &2.18            &0.53 &\textit{failed} &0.55 \\
Indoor\_45\_degree\_4   &\textit{failed} &1.53            &1.72 &9.79            &1.30 \\
Indoor\_45\_degree\_9   &\textit{failed} &0.49            &1.25 &4.74            &0.76 \\
\hline
Average                 &5.26            &2.10            &0.81 &7.26            &\textbf{0.70}  \\
\hline    

\hline        
        \end{tabular}
        \begin{tablenotes} 
        \item \textit{Unit:\%/m, 0.70 means the average error would be 0.70m for 100m motion.} 
        \end{tablenotes} 
        \end{threeparttable} 
        }
        \end{center}
        \vspace{-2.0em}
\end{table}

\begin{figure}[htb]  
    	\subfigtopskip=0pt 
    	\subfigbottomskip=0pt 
    	\subfigcapskip=-12pt 
        \centering
        \captionsetup{justification=justified}
        \subfigure[boxes\_translation]{
                \begin{minipage}[t]{1.0\columnwidth}
                \centering
                \includegraphics[width=1.0\columnwidth]{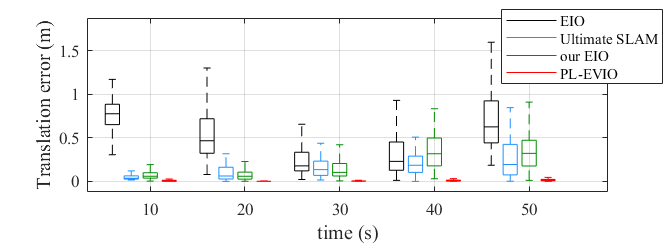}
                \label{boxplot_boxes_translation}
                \end{minipage}%
        }
        \subfigure[dynamic\_translation]{
                \begin{minipage}[t]{1.0\columnwidth}
                \centering
                \includegraphics[width=1.0\columnwidth]{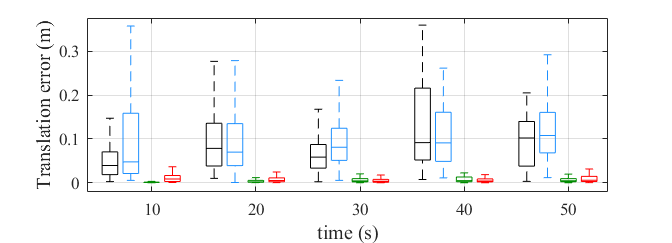}
                \label{boxplot_dynamic_translation}
                \end{minipage}%
        }
        \subfigure[poster\_6dof]{
                \begin{minipage}[t]{1.0\columnwidth}
                \centering
                \includegraphics[width=1.0\columnwidth]{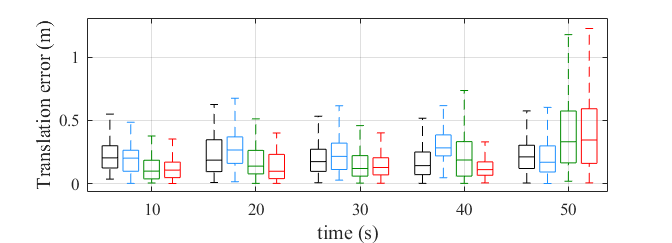}
                \label{boxplot_poster_6dof}
                \end{minipage}%
        }
        \caption{The relative pose error comparison of our PL-EVIO with EIO \cite{GWPHKU:ETH-EVIO}, Ultimate-SLAM \cite{GWPHKU:Ultimate-SLAM}, and our EIO \cite{GWPHKU:MyEVIO}}
        \label{boxplot}
        \vspace{-2.0em}
\end{figure}

Furthermore, we also evaluate our PL-EVIO with the other EIO works in publicly available Event Camera Datasets\cite{GWPHKU:event-camera-dataset_davis240c}, which is acquired by the DAVIS240C (240*180, event-sensor, image-sensor, IMU sensor).
It contains extremely fast 6-Dof motion and scenes with HDR.
We directly report the raw result in Ref.\cite{GWPHKU:Event-based-visual-inertial-odometry} \cite{GWPHKU:ETH-EVIO} \cite{GWPHKU:Ultimate-SLAM} \cite{HASTE-VIO} \cite{EKLT-VIO}, and \cite{GWPHKU:MyEVIO}.
As can be seen from Table \ref{DAVIS240C_Comparison}, our PL-EVIO achieves state-of-the-art performance.
Fig.\ref{boxplot} presents the relative error of our PL-EVIO against other methods, for the sequence \textit{box\_translation}, \textit{dynamic\_translation} and \textit{poster\_6dof}.
It's important to note that, although the Ultimate-SLAM\cite{GWPHKU:Ultimate-SLAM} (EVIO version) demonstrates performance similar to ours, it relies on different parameters for different sequence. 
While we consider parameter tuning to be impractical, we evaluate our methods using fixed parameters for various sequences during the evaluations.

\begin{table}[htbp]
        \begin{center}
        \caption{Accuracy Comparison of Our PL-EVIO with Other EIO/EVIO Works in DAVIS240c Dataset \cite{GWPHKU:event-camera-dataset_davis240c}}
        \label{DAVIS240C_Comparison}
        \resizebox{\columnwidth}{!}
        { 
        \begin{threeparttable}
        \renewcommand{\arraystretch}{1.0}
        \setlength{\tabcolsep}{1.0mm}
        \begin{tabular}{ccccccccc} 
        \hline  
        Sequence 
        & \makecell{Ref.\cite{GWPHKU:Event-based-visual-inertial-odometry} \\EIO} 
        & \makecell{Ref. \cite{GWPHKU:ETH-EVIO} \\EIO} 
        & \makecell{Ref. \cite{GWPHKU:Ultimate-SLAM}\\EIO}
        & \makecell{Ref. \cite{GWPHKU:Ultimate-SLAM}\\EVIO}
        & \makecell{Ref. \cite{HASTE-VIO}\\EIO}
        & \makecell{Ref. \cite{EKLT-VIO}\\EVIO}
        & \makecell{Our EIO \cite{GWPHKU:MyEVIO} \\EIO}
        & \makecell{Our PL-EVIO \\EVIO} \\
        \hline
        boxes\_translation      & 2.69 & 0.57 & 0.76             &0.27 & 2.55 & 0.48 & 0.34  & \textbf{0.06}\\
        hdr\_boxes              & 1.23 & 0.92 & 0.67             &0.37 & 1.75 & 0.46 & 0.40  & \textbf{0.10}\\
        boxes\_6dof             & 3.61 & 0.69 & 0.44             &0.30 & 2.03 & 0.84 & 0.61  & \textbf{0.21} \\
        dynamic\_translation    & 1.90 & 0.47 & 0.59             &\textbf{0.18} & 1.32 & 0.40 & 0.26  & 0.24\\
        dynamic\_6dof           & 4.07 & 0.54 & 0.38             &\textbf{0.19} & 0.52 & 0.79 & 0.43  & 0.48\\
        poster\_translation     & 0.94 & 0.89 & 0.15             &\textbf{0.12} & 1.34 & 0.35 & 0.40  & 0.54 \\
        hdr\_poster             & 2.63 & 0.59 & 0.49             &0.31 & 0.57 & 0.65 & 0.40  & \textbf{0.12}\\
        poster\_6dof            & 3.56 & 0.82 & 0.30             &0.28 & 1.50 & 0.35 & 0.26  & \textbf{0.14}\\
        \hline
        Average                 & 2.58 & 0.69 & 0.47             &0.25 & 1.45 & 0.54 & 0.39  & \textbf{0.24}  \\
        \hline       
        \end{tabular}
        \begin{tablenotes} 
        \item \textit{Unit:\%/m, 0.24 means the average error would be 0.24m for 100m motion.} 
        \end{tablenotes} 
        \end{threeparttable} 
        }
        \end{center}
        \vspace{-2.0em}
\end{table}

\subsection {Online Quadrotor-flight Evaluation} 
\label{Online Quadrotor-flight Evaluation}

\begin{figure}[htb]  
        \centering
        \captionsetup{justification=justified}
        \includegraphics[width=0.9\columnwidth]{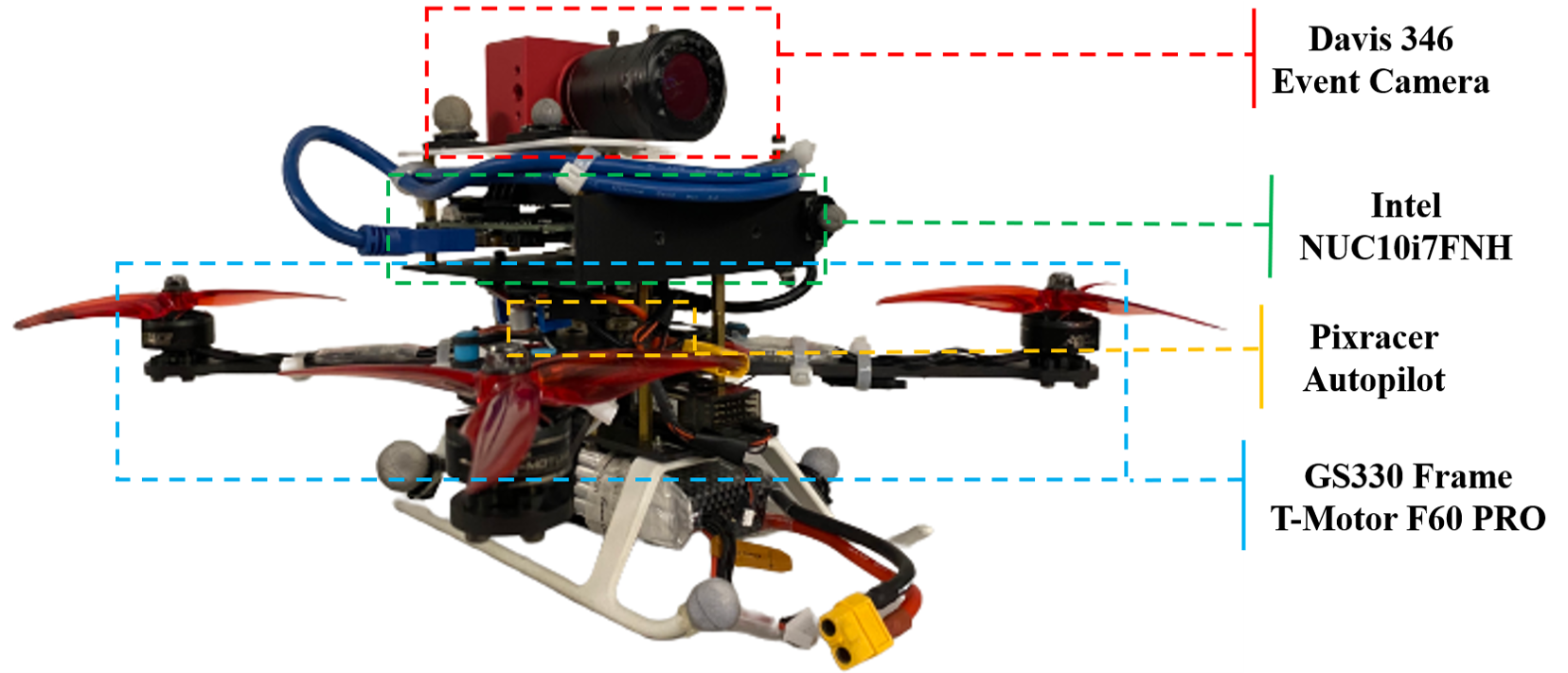}
        \caption{Our self-designed quadrotor platform. }  
        \label{quadrotor}
        \vspace{-1.0em}
\end{figure}

To further demonstrate the capabilities of our PL-EVIO, we perform real-world experiments on a self-designed quadrotor platform (shown in Fig.\ref{quadrotor}), carrying a forward-looking IniVation DAVIS346 sensor.
An Intel NUC10i7FNH computer running Ubuntu 20.04 is mounted on our quadrotor for onboard computational support. 
We use Pixracer (FMUv4) autopilot to run the PX4 flight stack. 
To alleviate disturbance from the motion capture system’s infrared light on the event camera, we add an infrared filter on the lens surface of the DAVIS346 camera. 
Note that the introduction of the infrared filter might cause the degradation of perception for both the event and image camera during the evaluation in subsection \ref{Evaluation in High-Dynamic-Range Scenarios}, \ref{Online Quadrotor-flight Evaluation}, \ref{Aggressive Quadrotor-flip Evaluation}, and \ref{Outdoor Large-scale Evaluation}.
The overall weight of our quadrotor is 1.364kg (GS330 frame with T-Motor F60). 

In the experiments, the reference trajectories are generated offline. The polynomial trajectory generation method \cite{mellinger2011minimum} is used to ensure the motion feasibility of the quadrotor.
To follow the generated trajectory, a cascaded feed-forward P.I.D. controller is constructed as a high-level position controller running on NUC. Given the position, velocity, and acceleration as inputs, the high-level feed-forward controller computes desired attitude and throttle sent to the low-level controller running on PX4.

We conduct four flight experiments to test the performance of autonomous trajectory tracking using our PL-EVIO.
The quadrotor is commanded to track different patterns as follows (\textit{Offboard} and \textit{Onboard} means using the VICON and our PL-EVIO as pose feedback control, respectively, while our PL-EVIO runs real-time and online calculations in the onboard computer):

\subsubsection{Offboard\_test\_1 and Onboard\_test\_1} 
\begin{figure}[htb]  
        \centering
        \captionsetup{justification=justified}
        \includegraphics[width=0.80\columnwidth]{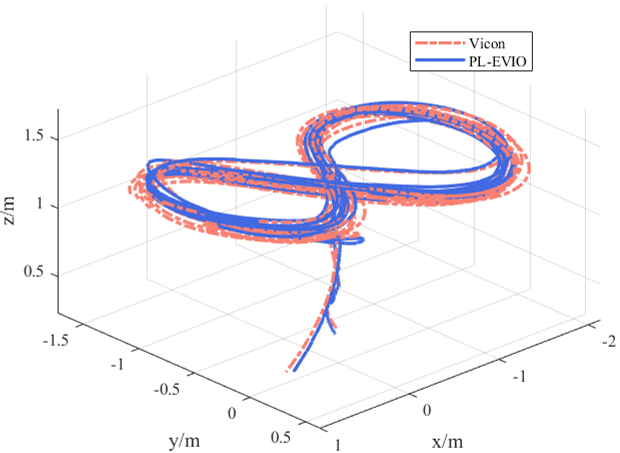}
        \caption{The estimated trajectory of our PL-EVIO on the quadrotor flight and its comparison against the ground truth (Taking the \textit{Onboard\_test\_1} as an example). }  
        \label{onboard_flight}
        \vspace{-1.0em}
\end{figure}

The states estimate from the VICON (\textit{Offboard\_test\_1}) and our PL-EVIO (\textit{Onboard\_test\_1}) are used for feedback control of the quadrotor which is commanded to track a figure-eight pattern with each circle being 0.625m in radius and 1.2m in height, shown in Fig.\ref{onboard_flight}.
The yaw angle of the commanded figure-eight pattern is fixed.
The quadrotor follows this trajectory ten times continuously during the experiment.
The 1000-HZ online calculation of our PL-EVIO is also recorded for accuracy comparison.

\begin{table}[htbp] 
        \renewcommand\arraystretch{1.2}
        \begin{center}
        \caption{Accuracy Comparison of Our PL-EVIO with Groundtruth in Quadrotor flight}
        \label{Real_drone_comparison}
        \resizebox{\columnwidth}{!}
        { 
        \begin{threeparttable}
        \tiny 
        \begin{tabular}{c|ccc|ccc} 
        \hline
        \multirow{2}*{Sequence}   &\multicolumn{3}{c|}{Translation Error} &\multicolumn{3}{c}{Rotation Error} \\
\cline{2-7}
~   & \makecell{Mean} 
    & \makecell{RMSE}
    & \makecell{Std} 
    & \makecell{Mean} 
    & \makecell{RMSE}
    & \makecell{Std}\\ 
\hline
Offboard\_test\_1      & 0.054 & 0.061 &  0.028 & 0.094 & 0.095 & 0.015            \\
Onboard\_test\_1       & 0.078 & 0.084 & 0.030  & 0.078  & 0.087 & 0.039              \\
Onboard\_test\_2       & 0.081 & 0.093 & 0.046  & 0.056  & 0.059 & 0.019           \\
\hline    
        \end{tabular}
        \begin{tablenotes} 
        \item \textit{Unit: m for translation and deg for rotation} 
        \end{tablenotes}
        \end{threeparttable} 
        }
        \end{center}
\end{table}

\subsubsection{Onboard\_test\_2} 
\begin{figure}[htb]  
        \centering
        \includegraphics[width=0.95\columnwidth]{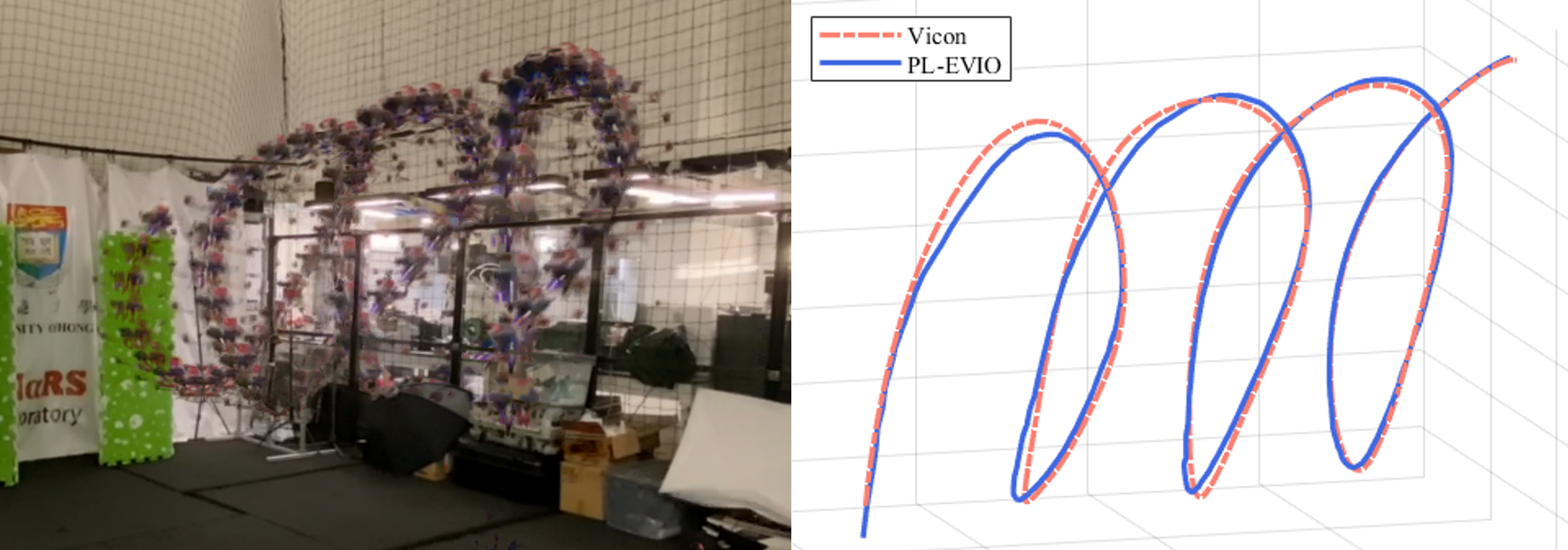}
        \caption{Onboard quadrotor flight in screw pattern using our PL-EVIO as feedback control}  
        \label{screw}
        \vspace{-2.0em}
\end{figure}

\begin{figure*}[htb]  
    	\subfigtopskip=0pt 
    	\subfigbottomskip=0pt 
    	\subfigcapskip=-3pt 
        \centering
        \subfigure[X-axis]{
                \begin{minipage}[t]{0.8\columnwidth}
                \centering
                \includegraphics[width=0.8\columnwidth]{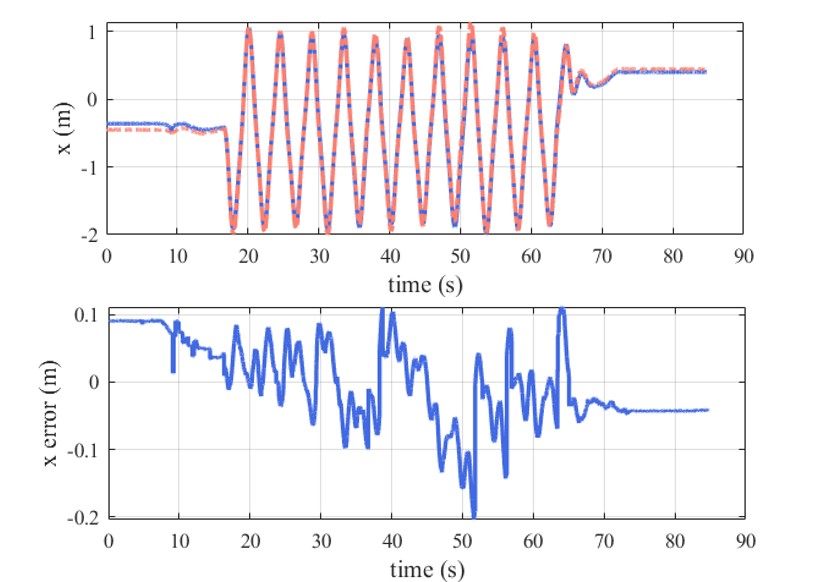}
                \label{online_x}
                \end{minipage}%
        }\hspace{-20mm}
        \subfigure[Y-axis]{
                \begin{minipage}[t]{0.8\columnwidth}
                \centering
                \includegraphics[width=0.8\columnwidth]{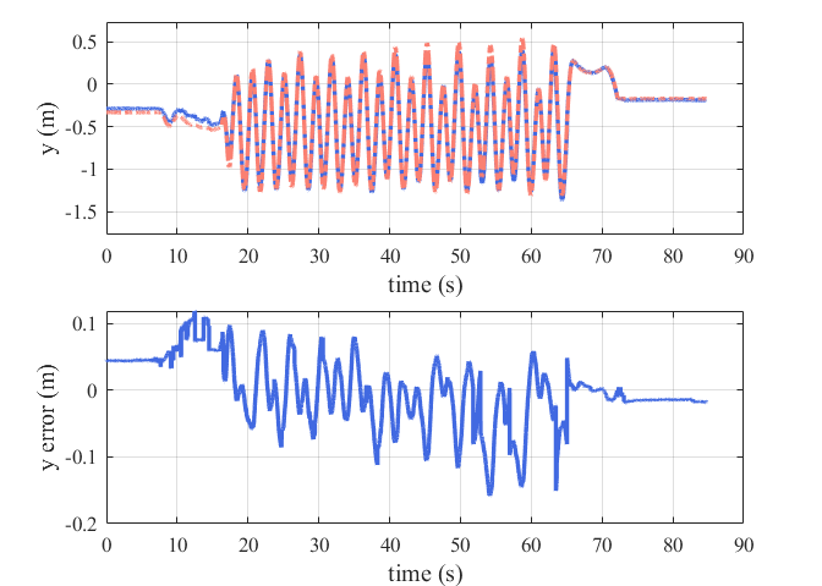}
                \label{online_y}
                \end{minipage}%
        }\hspace{-20mm}
        \subfigure[Z-axis]{
                \begin{minipage}[t]{0.8\columnwidth}
                \centering
                \includegraphics[width=0.8\columnwidth]{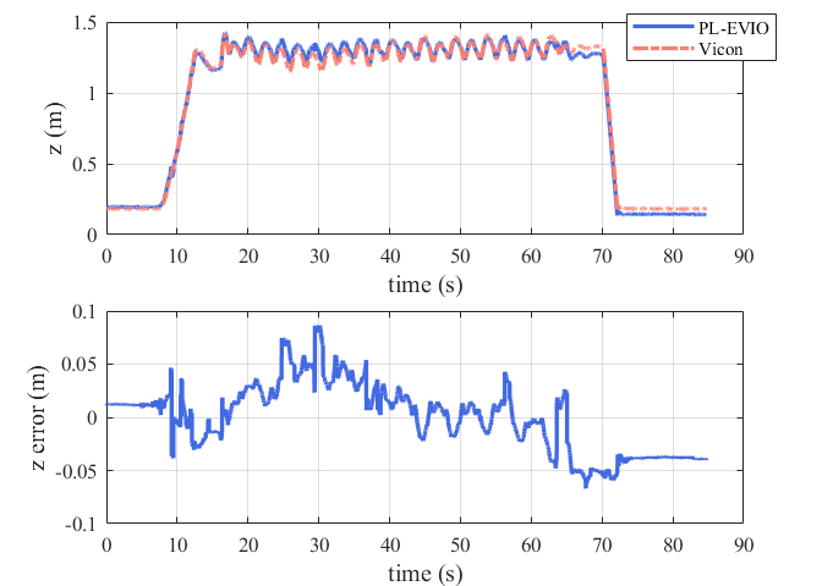}
                \label{online_z}
                \end{minipage}%
        }
       \subfigure[Roll-axis]{
                \begin{minipage}[t]{0.8\columnwidth}
                \centering
                \includegraphics[width=0.8\columnwidth]{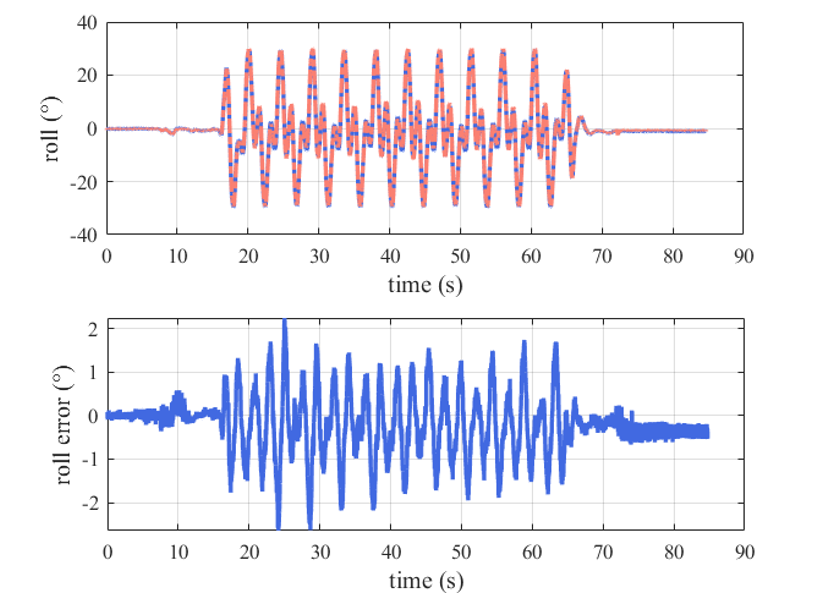}
                \label{online_roll}
                \end{minipage}%
        }\hspace{-20mm}
        \subfigure[Pitch-axis]{
                \begin{minipage}[t]{0.8\columnwidth}
                \centering
                \includegraphics[width=0.8\columnwidth]{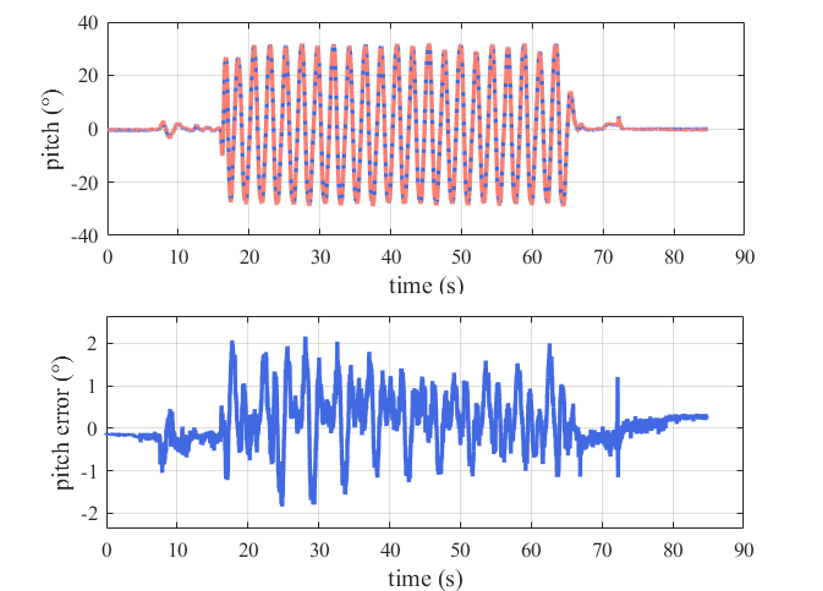}
                \label{online_pitch}
                \end{minipage}%
        }\hspace{-20mm}
        \subfigure[Yaw-axis]{
                \begin{minipage}[t]{0.8\columnwidth}
                \centering
                \includegraphics[width=0.8\columnwidth]{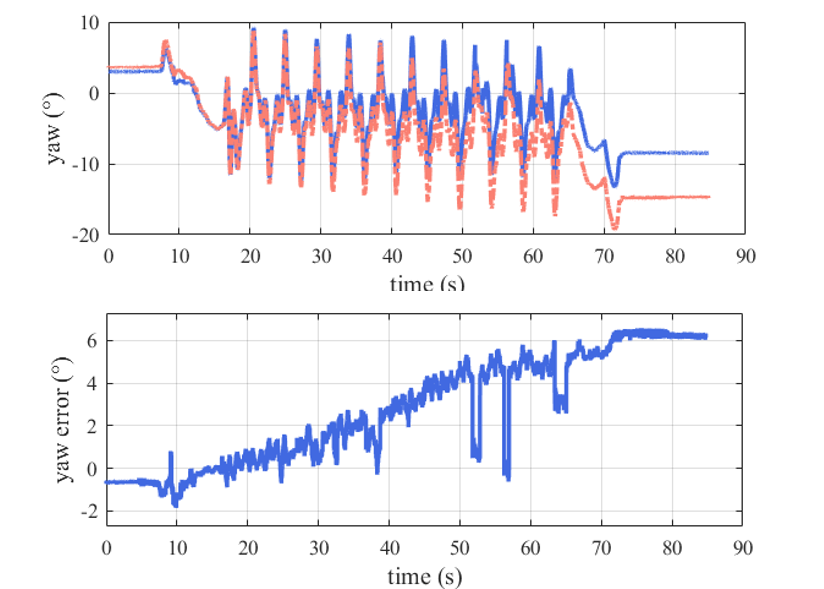}
                \label{online_yaw}
                \end{minipage}%
        }
        \caption{The position, orientation, and the corresponding errors of our PL-EVIO in onboard flight compared with the ground truth from VICON (Taking the \textit{Onboard\_test\_1} as example).}
        \label{online_fix_eight}
        \vspace{-1.0em}
\end{figure*}

The states estimate from our PL-EVIO are used for feedback control of the quadrotor which is commanded to track a screw pattern shown in Fig.\ref{screw}.
The quadrotor follows this trajectory ten times continuously during the experiment.
The 1000-HZ onboard state estimates of our PL-EVIO enable real-time feedback control of the quadrotor.
The ground truth is obtained from VICON. 
The translation and rotation error are shown in Table \ref{Real_drone_comparison}.
Taking the \textit{Onboard\_test\_1} as an example, in Fig.\ref{onboard_flight} and Fig.\ref{online_fix_eight}, we further illustrate the estimated trajectories (translation and rotation) of our PL-EVIO against the ground truth, as well as their corresponding errors. 
The total trajectory length is 101.15m. 
The translation errors in the X, Y, and Z dimensions are all within 0.1m, while the rotation error of the Roll and Pitch dimensions are within 2\degree, and the one in the Yaw dimension is within 6\degree.

\subsection {Aggressive Quadrotor-flip Evaluation} 
\label{Aggressive Quadrotor-flip Evaluation}

\begin{figure}[H]  
        \centering
        \includegraphics[width=0.95\columnwidth]{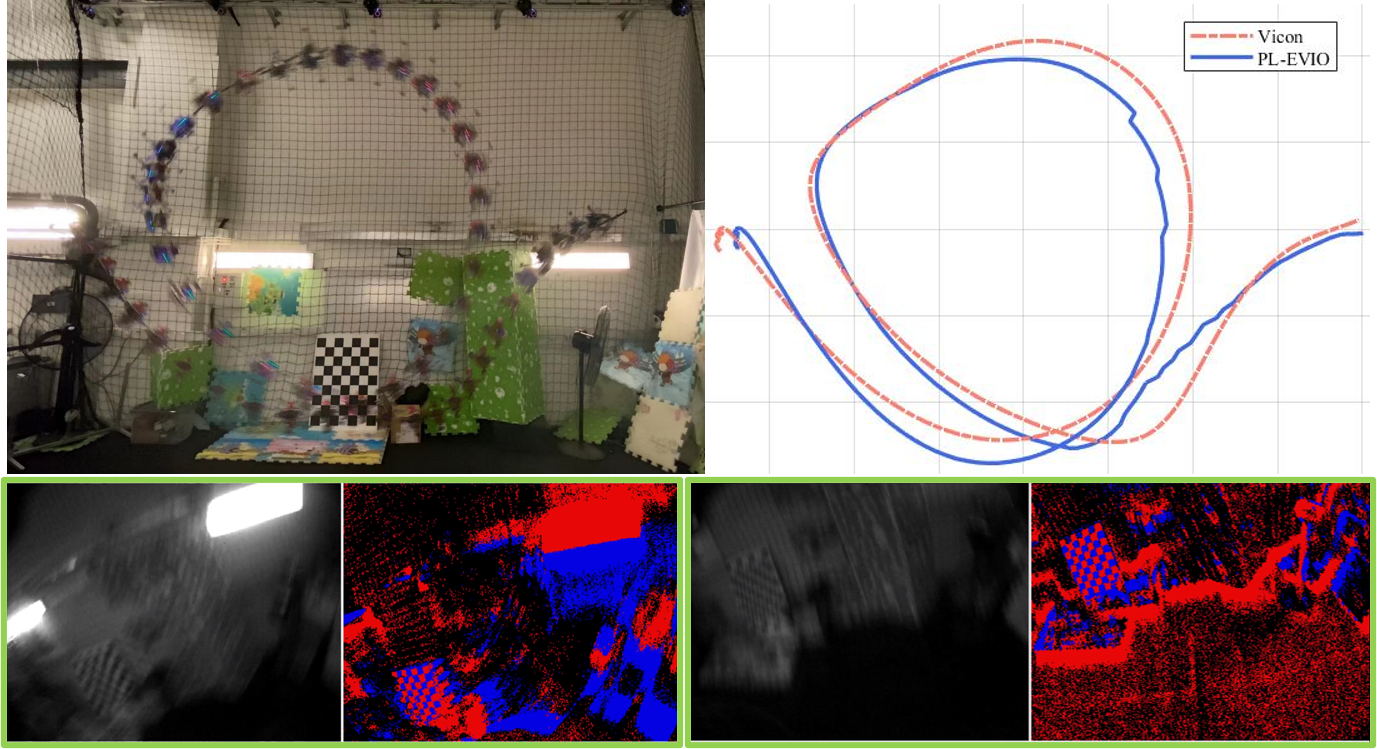}
        \caption{The estimated trajectory of our PL-EVIO on the quadrotor flip, and image/event measurement.}  
        \label{quadrotor-flip}
\end{figure}

In this section, we further conduct onboard quadrotor flip experiments to evaluate the performance of our PL-EVIO in aggressive motion.
The estimated trajectory of our PL-EVIO compared with the ground truth from VICON during the flip evaluations can be seen in Fig.\ref{quadrotor-flip}. 
The total length of the trajectory is 15m. The mean translation error and the mean angular error are 0.097m and 6.0\degree, respectively.
Despite the extreme velocity of the motion, our PL-EVIO successfully tracks the quadrotor pose with high accuracy.
Note that our PL-EVIO is run onboard during the quadrotor flip experiments. 
There are only a few image measurements captured during aggressive motion due to motion blur, whereas the event measurements are severely limited when the quadrotor is hovering.
Thanks to our well-designed feature management and the complementarity of three kinds of features, our PL-EVIO can provide robust and good performance in multiple quadrotor flip experiments.

\subsection {Real-time Analysis} 
\label{Real-time Performance}
We assessed the real-time performance of our system on quadrotor flight using an Intel NUC10i7FNH as the computing platform. 
The computational allocations are presented in Table \ref{real-time-performance}.
The proposed algorithm sequentially processes the event queue, with the event front-end completed within 9 ms, the image front-end completed within 3 ms, and overall optimization completed within 50 ms, without any hardware acceleration.
In order to achieve low latency, we employ a loosely-coupled approach to directly propagate the latest EVIO estimation along with the IMU measurements.
This results in IMU-rate EVIO outputs that can reach up to 1000 Hz.
This is critical for achieving onboard quadrotor flight, using our PL-EVIO as pose feedback control, as discussed in Sections \ref{Online Quadrotor-flight Evaluation} and \ref{Aggressive Quadrotor-flip Evaluation}.
Due to the reliable, low-drift, and low-latency characteristics of our PL-EVIO, the flight control system can quickly obtain accurate pose feedback, thereby ensuring the success of onboard quadrotor flight.
To further demonstrate the real-time capabilities of our proposed PL-EVIO system, which offers onboard pose feedback for quadrotor flights, we encourage readers to refer to the video demos\footnote{\url{https://b23.tv/VqGMkyD}}.

\begin{table}[htbp]
\begin{center}
\caption{Time Consumption of Different Modules in Our PL-EVIO }
\label{real-time-performance}
\resizebox{\columnwidth}{!}
{ 
    \begin{threeparttable}
    \tiny 
    \begin{tabular}{c|c|c} 
        \hline
        \multicolumn{2}{c}{Modules} & Time-cost (ms) \\
        \hline
        \multirow{5}*{Event Front-end} & Point-based event feature detection & 0.41 \\
                                       & Point-based event feature tracking  & 0.62 \\
                                       & Line-based event feature detection  & 2.88  \\
                                       & Line-based event feature Matching   & 3.57  \\
                                       & Total                               & 8.68  \\
        \hline  
        \multirow{3}*{Image Front-end} & Point-based image feature detection & 0.70 \\
                                       & Point-based image feature tracking  & 0.42 \\
                                       & Total                               & 2.29  \\ 
        \hline
        \multirow{7}*{Back-end}        & Construct point-based event residual  & 0.089 \\
                                       & Construct line-based event residual   & 0.0026 \\
                                       & Construct point-based image residual  & 0.66 \\
                                       & Construct marginalization residuals   & 7.50  \\
                                       & Solve graph optimization using Ceres  & 37.66  \\
                                       & IMU forward                           & 0.0056 \\
                                       & Total                                 & 50.49  \\ 
        \hline    
    \end{tabular}
    \end{threeparttable} 
}
\end{center}
\vspace{-3.0em}
\end{table}

\subsection {Outdoor Large-scale Evaluation} 
\label{Outdoor Large-scale Evaluation}

\begin{figure*}[htbp]  
        \subfigbottomskip=0pt 
        \subfigcapskip=-3pt 
        \centering
        \captionsetup{justification=justified}
        \subfigure[HKU\_centennial\_garden]{
            \begin{minipage}[t]{2.0\columnwidth}
            \centering
            \includegraphics[width=1.0\columnwidth]{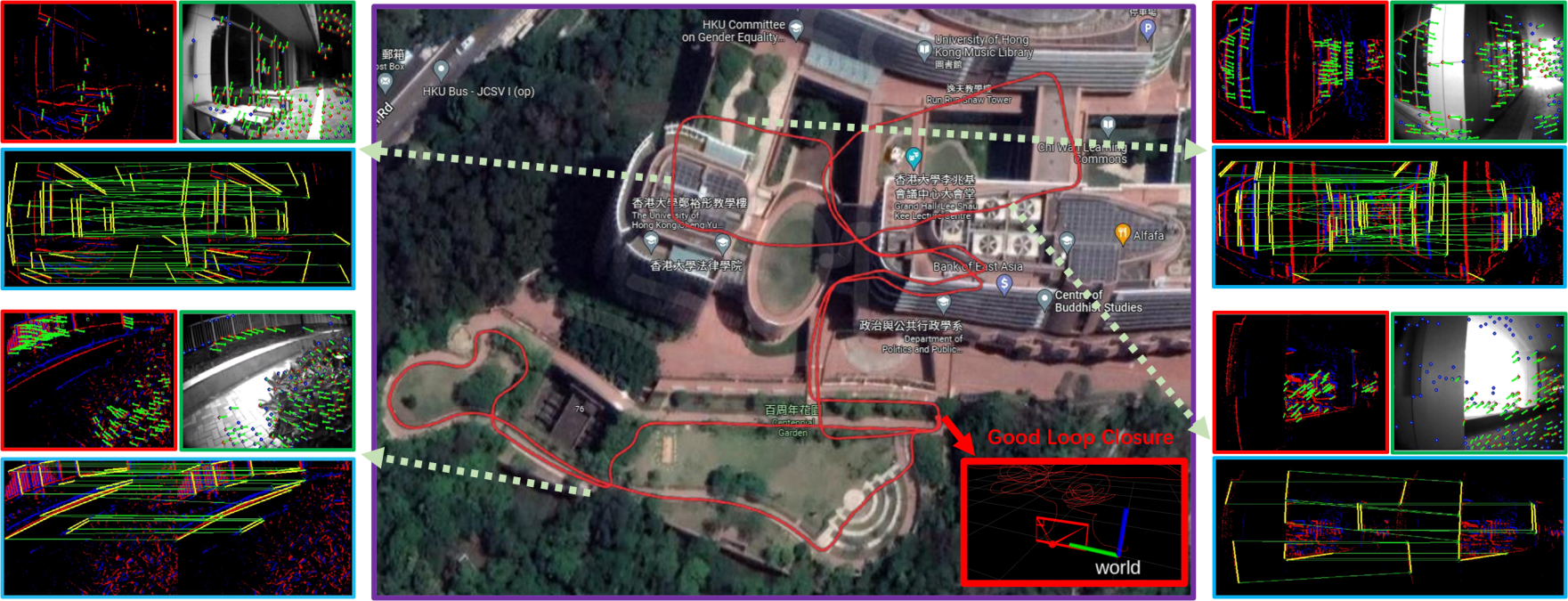}
            \label{Nature-scenarios}
            \end{minipage}%
        }
        \subfigure[HKU\_main\_building]{
            \centering
            \includegraphics[width=2.0\columnwidth]{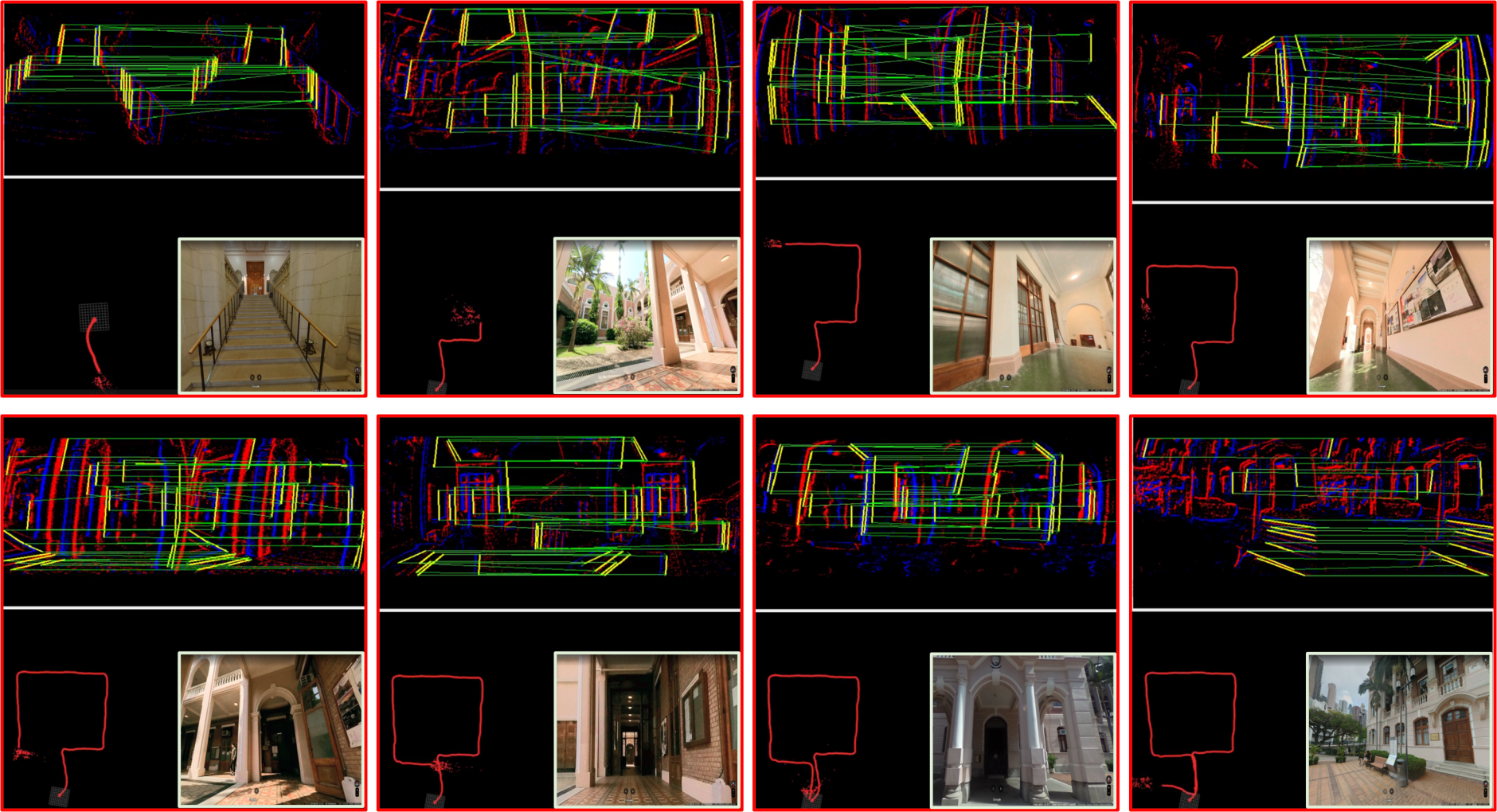}
            \label{Human-made-scenarios}
        }
        \caption{
        (a)The estimated trajectory of our PL-EVIO in the outdoor environment. 
        We also visualize the detection and tracking situation of the event-corner features, line-based event features, and point-based image features, during the experiment.
        The combination of these features provides more structures and constraints in the scene that ensure robustness.
        (b) The estimated trajectory of our PL-EVIO as well as the detection and matching performance of the line-based event features.}
        \label{outdoor_evaluation}
        \vspace{-1.5em}
\end{figure*}%

\subsubsection{Natural Scenarios}

In this section, we evaluate our PL-EVIO system in a large-scale environment that encompasses the HKU campus.
This environment includes features such as moving pedestrians, low-texture areas, long-term movement, strong sunlight, and indoor \& outdoor transitions. 
We also return to the same location after a large loop to evaluate the loop closure.
The total evaluation length is approximately 980 m, covering an area of approximately 160 m in length, 100 m in width, and 10 m in height changes.
The estimated trajectory is aligned with the Google map and can be seen in Fig.\ref{Nature-scenarios}.
The results show that our PL-EVIO performed almost drift-free in this long-term motion evaluation.
The complementarity of three different kinds of features (e.g. line-based event features for human-made environment, point-based event features for HDR scene, and point-based image features for good lighting scene) ensure the robust and reliable state estimation.

\subsubsection{Human-made Scenarios}


We further conduct additional evaluations specifically focusing on building scenarios. 
Utilizing the line features can better represent the geometric information constraints in Human-made structures, as illustrated in Fig.\ref{Human-made-scenarios}.
The experimental results demonstrate that after a long-distance loop of approximately 260 m within the interior of the building, our PL-EVIO system maintains high accuracy, forming a complete square shape without significant drifts. 
To assess this accuracy, we specifically choose the gate of the building as the starting and ending point for quantitative evaluation, and the end-to-end distance showed an error of 0.61 m.
Owing to the additional geometric structural information provided by our proposed event-based line features, our PL-EVIO achieves low drift and reliable performance in this large-scale environment.

\section{CONCLUSIONS} 
\label{CONCLUSIONS}
In this paper, we propose a robust, highly-accurate, and real-time optimization-based monocular VIO that tightly fuses the event, image, and IMU together, with point and line features.
The combination of point-based event-corner features, line-based event features, and point-based image features would provide more geometric constraints on the structure of the environment.
Finally, we show superior performance by comparing against other state-of-the-art open-source image-based or event-based VIO implementations in different challenge datasets.
Meanwhile, through extensive experiments including extremely aggressive motion and large-scale evaluation, we also show that our PL-EVIO pipeline is able to leverage the properties of the standard camera and the event camera with different features to provide robust state estimation.
We hope that this work can inspire other researchers and industries to push wide applications for event cameras on robotics and perception. 
In our future work, event-based multi-sensor fusion, including a wider range of local perception (such as LiDAR), and global perception (such as visible light positioning \cite{GWPHKU:My2021TIM} for indoor, or GPS for outdoor), might be deeply studied to exploit the complementary advantage of different sensors with event cameras.

\begin{appendices}
\setcounter{table}{0} 
\setcounter{figure}{0} 
\setcounter{equation}{0} 
\renewcommand{\thetable}{\thesection-\Roman{table}} 
\renewcommand{\theequation}{\thesection-\arabic{equation}} 
\renewcommand{\thefigure}{\thesection-\arabic{figure}} 

\section{Ablation Study}
\label{Ablation experiment}
\begin{figure*}[htb]  
        \subfigbottomskip=0pt 
        \subfigcapskip=1.0pt 
        \centering
        \captionsetup{justification=justified}
        \subfigure[Raw Event Stream]{
            \begin{minipage}[t]{1.0\columnwidth}
            \centering
            \includegraphics[width=1.0\columnwidth]{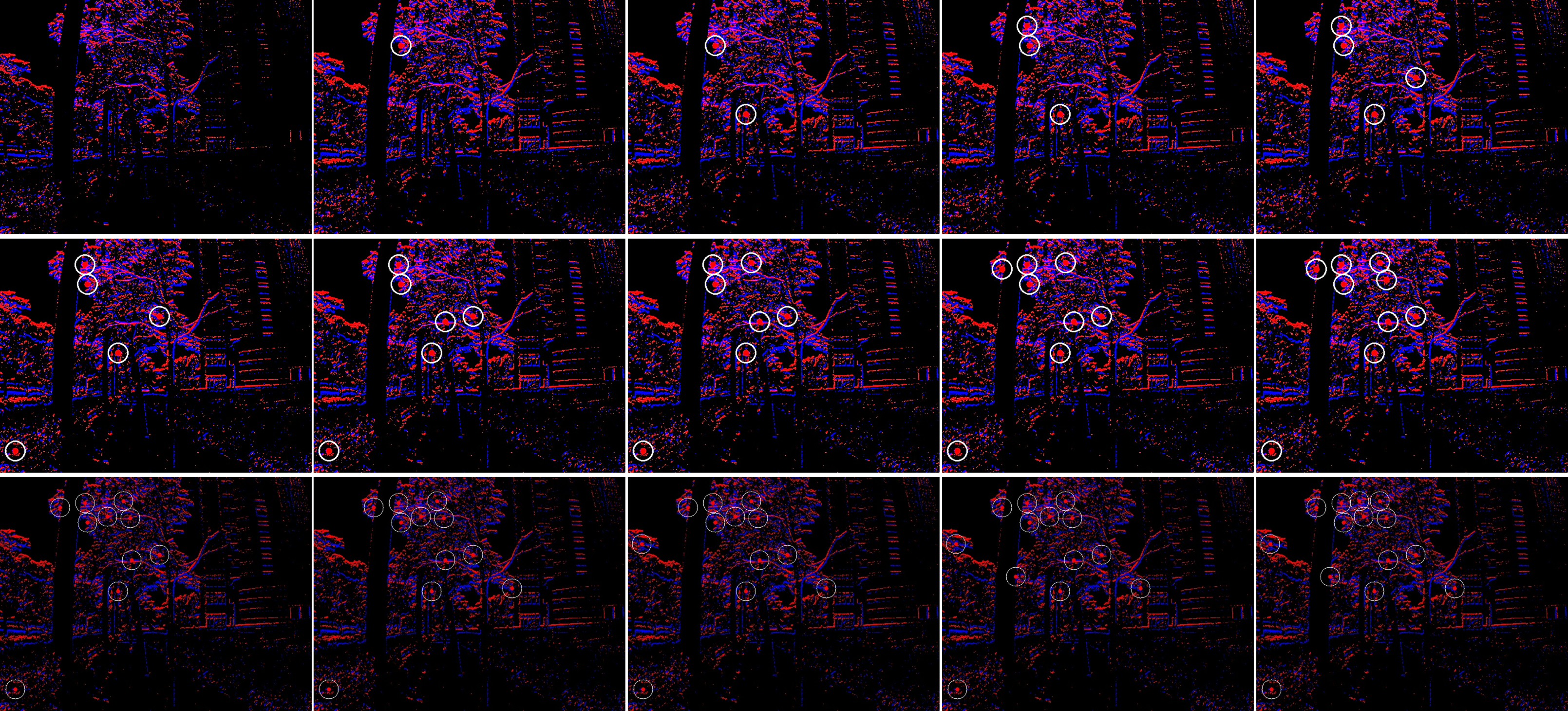}
            \end{minipage}%
        }
        \subfigure[TS with Polarity]{
            \begin{minipage}[t]{1.0\columnwidth}
            \centering
            \includegraphics[width=1.0\columnwidth]{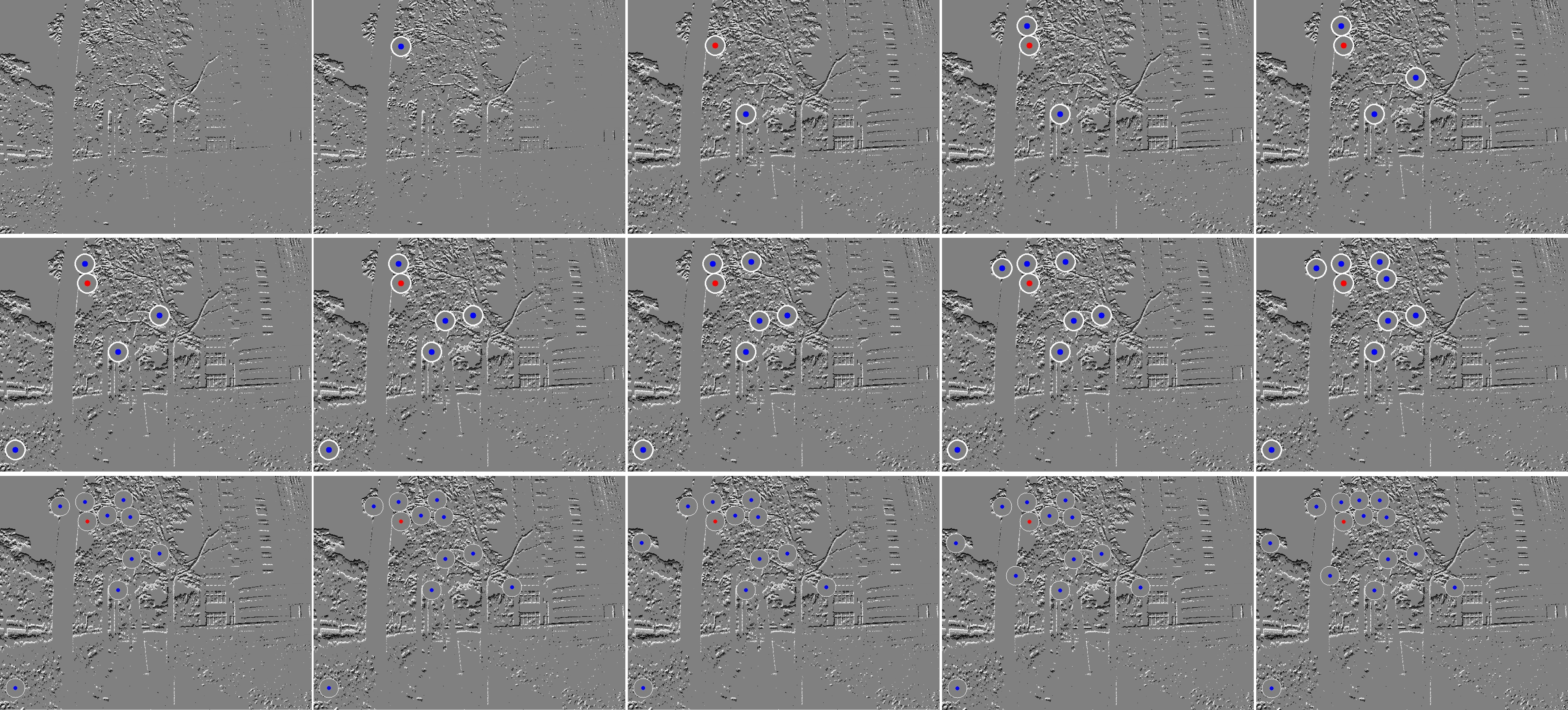}
            \end{minipage}%
        }
        \caption{
        Event-corner features generation.
        The event-corner features are firstly extracted from the asynchronous event stream (a), then the TS with polarity (b) is used as a mask to further select the event-corner features, ensuring a uniform distribution.
        }  
        \label{event-coner-generation}
\end{figure*}

Our approach involves extracting event-corner features from events-only data and line-based features from the event mat.
These two types of features are then associated using a spatio-temporal locality scheme based on exponential decay, which is commonly referred to as TS.
The TS was converted from SAE with the exponential decay kernel.
To enhance the accuracy of event-based point and line tracking, we incorporate polarity into the TS ($T_{p}(\boldsymbol{x},t)$), which can be represented as follows:
\begin{equation}
T_{p}(\boldsymbol{x},t)=p \cdot \exp(-\frac{t-t_{last}(\boldsymbol{x})}{\eta}) 
\end{equation}
We use the LK optical flow on the $T_{p}(\boldsymbol{x},t)$ to associate the current event-corner with the most recent event-corner in the kernel operation, assuming that it is the same event-corner in a recent position.
Our approach utilizes the motion variance characteristics of the $T_{p}(\boldsymbol{x},t)$ to retain relevant context while associating event-based point and line features into tracks to ensure computational efficiency in the front-end.
In our previous work \cite{GWPHKU:MyEVIO}, we have presented the process of generating uniform event-corner features (as shown in Fig.\ref{event-coner-generation}) and discussed the rationale for employing our $T_{p}(\boldsymbol{x},t)$ for tracking event-corner features.
Additionally, we also provided the normalized TS without polarity ($T_{np}(\boldsymbol{x},t)$):
\begin{equation}
T_{np}(\boldsymbol{x},t)=(\frac{255.0}{\max(T^{'})-\min(T^{'})}) \cdot (T^{'}-\min(T^{'}))
\end{equation}
where, $T^{'}$ can be obtained from Eq.\ref{TS_p}.

\subsection{Ablation Study on Different Event Representations for Event-based Feature Tracking}
\begin{figure*}[htb]  
        \centering
        \captionsetup{justification=justified}
        \includegraphics[width=2.0\columnwidth]{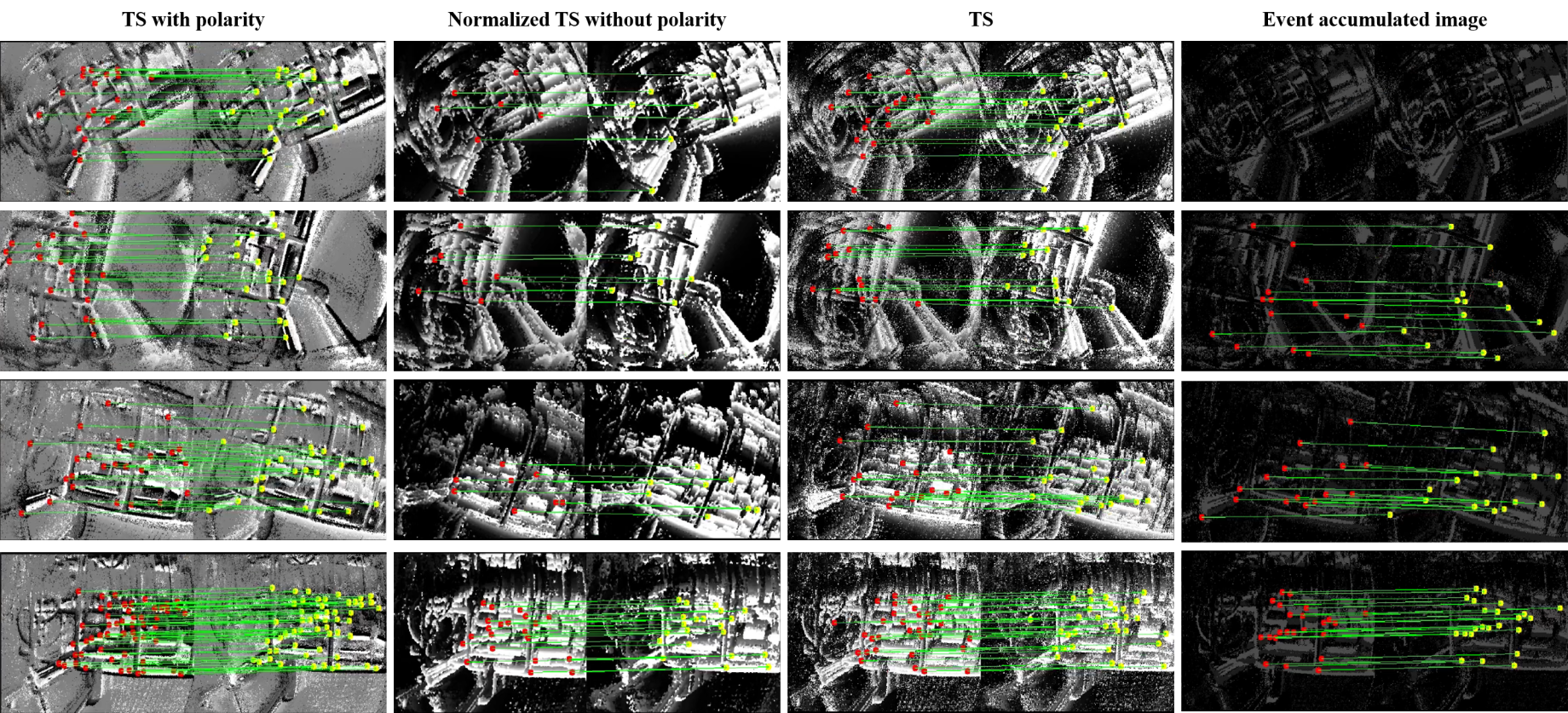}
        \caption{
        The performance of the event-corner features tracking in different event representations. 
        Note that the VIO is a highly nonlinear system, it is hard to prove the performance through a single timestamp.
        Therefore we refer the readers to our video demo:\url{https://b23.tv/eIRQMST}, which shows the reliable performance of our PL-EVIO.
        }  
        \label{ablation_study_tracking}
\end{figure*}

In this section, we focus on conducting an ablation study of the event-based point tracking performance using different event representations, including our $T_{p}(\boldsymbol{x},t)$, $T_{np}(\boldsymbol{x},t)$, TS in \cite{GWPHKU:ESVO}, and the event accumulated image in \cite{GWPHKU:EVO} and \cite{GWPHKU:Ultimate-SLAM}.
It should be noted that we previously only used $T_{np}(\boldsymbol{x},t)$ for loop closure detection, while we merely investigate its feature tracking performance in front-end for this ablation study.
During the ablation experiments, we employ our PL-EIO framework to control variables by only altering different event representations used for event-based feature tracking, while keeping the event feature points generated from asynchronous event streams unchanged.
The qualitative results are presented in Fig. \ref{ablation_study_tracking}, and the quantitative evaluations are reported in the "Event Representations" section of Table \ref{ablation_study_table}. 
We utilize absolute trajectory error (ATE) aligning the estimated trajectory with ground truth using 6-DOF transformation (in SE3) to quantitatively evaluate the accuracy.
The results indicate that only our proposed $T_{p}(\boldsymbol{x},t)$ is capable of reliably estimating the state, while other event representations used for event-based feature tracking failed in the challenge situation.
This could be due to the insufficient intensity information available to satisfy the requirements of LK optical flow for event-based feature tracking.
For example, the image generated from event streams\cite{GWPHKU:Ultimate-SLAM}\cite{GWPHKU:EVO} is a binary edge image consisting of only two possible pixel values (0 or 1), which lacks the necessary information for accurately calculating gradients. 
Consequently, it becomes difficult to determine the direction and magnitude of motion of feature points, resulting in increased difficulty for optical flow to remove local outliers.
In contrast, our TS with polarity can ensure reliable data association between event features of adjacent frames, which effectively prevents miss-matches.

\begin{table}[htbp]
        \vspace{-0.8em}
        \begin{center}
        \caption{Accuracy Result of the Ablation Study in HKU\_agg\_flip}
        \label{ablation_study_table}
        \resizebox{\columnwidth}{!}
        { 
        \begin{threeparttable}
        \renewcommand{\arraystretch}{1.0}
        \setlength{\tabcolsep}{1.0mm}
        \begin{tabular}{ccccc} 
        \hline  
        Event Representations                           & $T_{p}(\boldsymbol{x},t)$      & $T_{np}(\boldsymbol{x},t)$                           & TS         & Event-image \\
        \hline
        \makecell{PL-EIO\\Event+IMU}                    & 0.25              & \textit{failed}                         & \textit{failed}         & \textit{failed} \\
        \hline
        \hline
        Time Decay Kernel                               & 10      & 20                                       & 60                     & 100 \\
        \hline
        \makecell{PL-EIO\\Event+IMU}                    & 0.30    & 0.25                                     & \textit{failed}        & \textit{failed} \\
        \hline
        \hline
        \multirow{2}*{Methods}    &\makecell{PL-EVIO\\Event+Image+IMU}   & \makecell{Ref.\cite{GWPHKU:Ultimate-SLAM}\\Event+Image+IMU}  & \makecell{ Ref.\cite{GWPHKU:EVO}\\Event} & \makecell{Ref.\cite{GWPHKU:ESVO}\\Stereo Event} \\
        \cline{2-5}
        ~                                               & 0.12     & 2.66                                     & \textit{failed}         & \textit{failed}      \\
        \hline       
        \end{tabular}
        \end{threeparttable} 
        }
        \end{center}
        \vspace{-2.0em}
\end{table}

\subsection{Ablation Study on Time Decay Kernels of the TS with Polarity}
\begin{figure*}[htb]  
        \subfigbottomskip=-1pt 
        \subfigcapskip=-10pt 
        \centering
        \captionsetup{justification=justified}
        \subfigure[HKU\_centennial]{
            \begin{minipage}[t]{2.0\columnwidth}
            \centering
            \includegraphics[width=1.0\columnwidth]{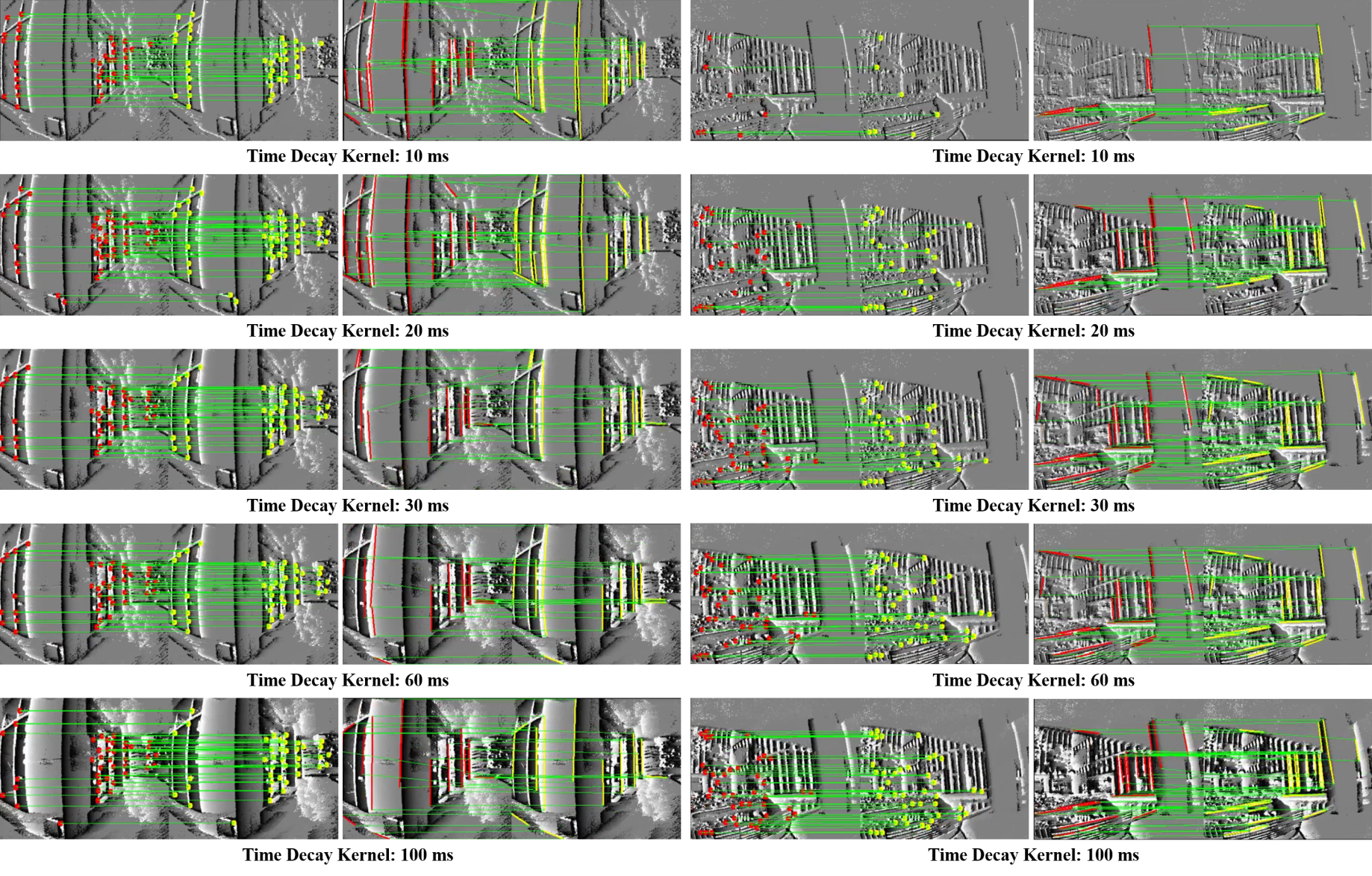}
            \label{ablation_study:outdoor}
            \end{minipage}%
        }
        
        \subfigure[HKU\_agg\_flip]{
            \begin{minipage}[t]{2.0\columnwidth}
            \centering
            \includegraphics[width=1.0\columnwidth]{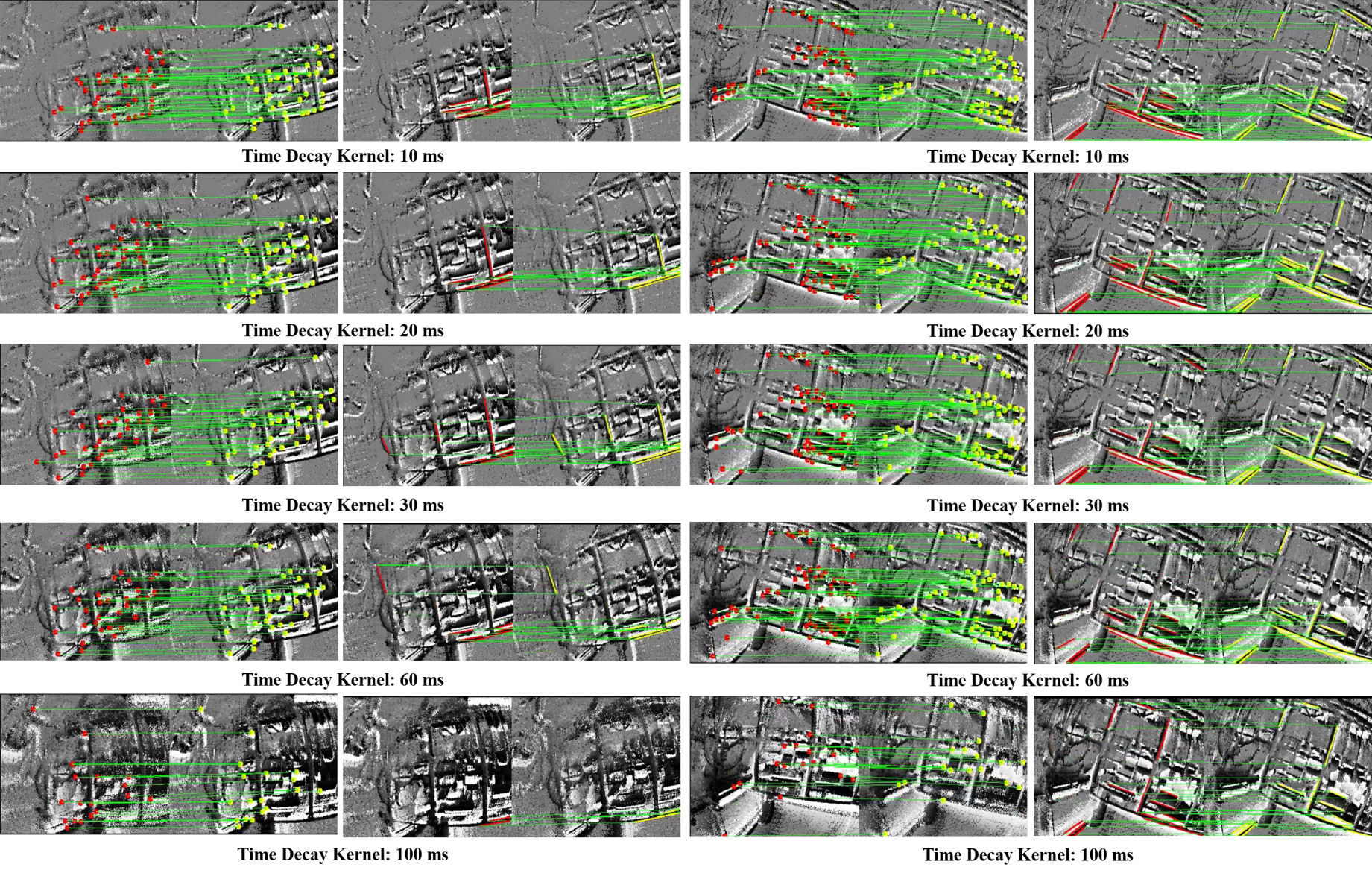}
            \label{ablation_study:indoor}
            \end{minipage}%
        }
        \caption{
        The tracking and matching performance of event-based point and line features in various time decay parameters. 
        We only evaluate the tracking and matching performance in (a) outdoor environments with large-scale and (b) indoor environments with aggressive motion, respectively.
        }  
        \label{ablation_study}
\end{figure*}

In order to further investigate the impact of time decay kernels on our event representations ($T_{p}(\boldsymbol{x},t)$) used for event feature tracking, we conducted ablation experiments on the time decay kernels. 
The qualitative results on the event-based point and line features are shown in Fig.\ref{ablation_study}.
From Fig.\ref{ablation_study}(a), we can observe that the tracking and matching performance of event-based point features and line features are not significantly different across various exponential decay kernels. 
This may be attributed to the normal texture conditions and fewer triggered events at a far distance in outdoor environments. 
However, upon careful observation, we still can notice that larger time decay kernels result in coarser edge contours in the edge regions, which may introduce systematic errors to the visual front-end. 
In contrast, in indoor environments (as shown in Fig.\ref{ablation_study}(b)) with abundant texture, we found that the larger time decay kernels lead to a significant decrease in the successful matching of both event-based line and point features.
This might be due to the trailing effect caused by a large time decay kernel, which can create negative effects (i.e. lead to failure during aggressive motion) similar to motion blur.
Therefore, we choose a time decay kernel of 20ms to ensure sufficient information for event-based feature matching and tracking while minimizing blur and trailing effects. 
We quantitatively evaluate the influence of the time decay kernel on the performance of pose estimation through the \textit{Time Decay Kernel} part of Table \ref{ablation_study_table}. 
Furthermore, we also compare the performance of our PL-EVIO using $T_{p}(\boldsymbol{x},t)$ as event representation with other event representations from Ref.\cite{GWPHKU:Ultimate-SLAM}, \cite{GWPHKU:EVO}, and \cite{GWPHKU:ESVO}, in the \textit{Methods} part of Table \ref{ablation_study_table}.

\end{appendices}


\bibliographystyle{IEEEtran} 
\bibliography{references.bib} 

\clearpage 
\begin{IEEEbiography}[{\includegraphics[width=1in,height=1.25in,clip,keepaspectratio]{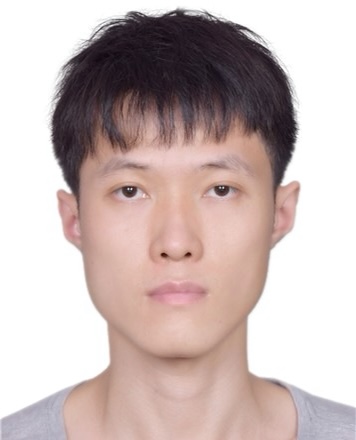}}]{Guan Weipeng}
obtained his Bachelor's and Master's degrees from the South China University of Technology. 
He is currently pursuing his PhD degree at the University of Hong Kong. 
He has worked with several reputable organizations, including: Samsung Electronics, Huawei Technologies, TP-LINK, The Chinese Academy of Sciences, The Chinese University of Hong Kong, The Hong Kong University of Science and Technology, etc.  
He has also served as a technical consultant for multiple companies, such as TCL.
Moreover, he has authored or co-authored over 60 research articles in prestigious international journals and conferences, as well as holds more than 40 authorized patents. 
His research interests primarily focus on robotics, event-based VO/VIO/SLAM, visible light positioning, etc.
\end{IEEEbiography}

\begin{IEEEbiography}[{\includegraphics[width=1in,height=1.25in,clip,keepaspectratio]{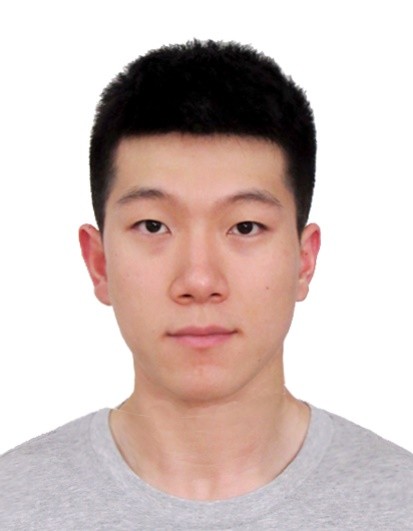}}]{Chen Peiyu}
received the BSc degrees in automation from the Nanjing University of Science and Technology, China, and MSc degrees in computer control \& automation from the Nanyang Technological University, Singapore, in 2020 and 2022, respectively. 
He is currently working forward the Ph.D. degree at the University of Hong Kong.  
His research interests include robotics, visual-inertial simultaneous localization and mapping, nonlinear control, and so on.
\end{IEEEbiography}

\begin{IEEEbiography}[{\includegraphics[width=1in,height=1.25in,clip,keepaspectratio]{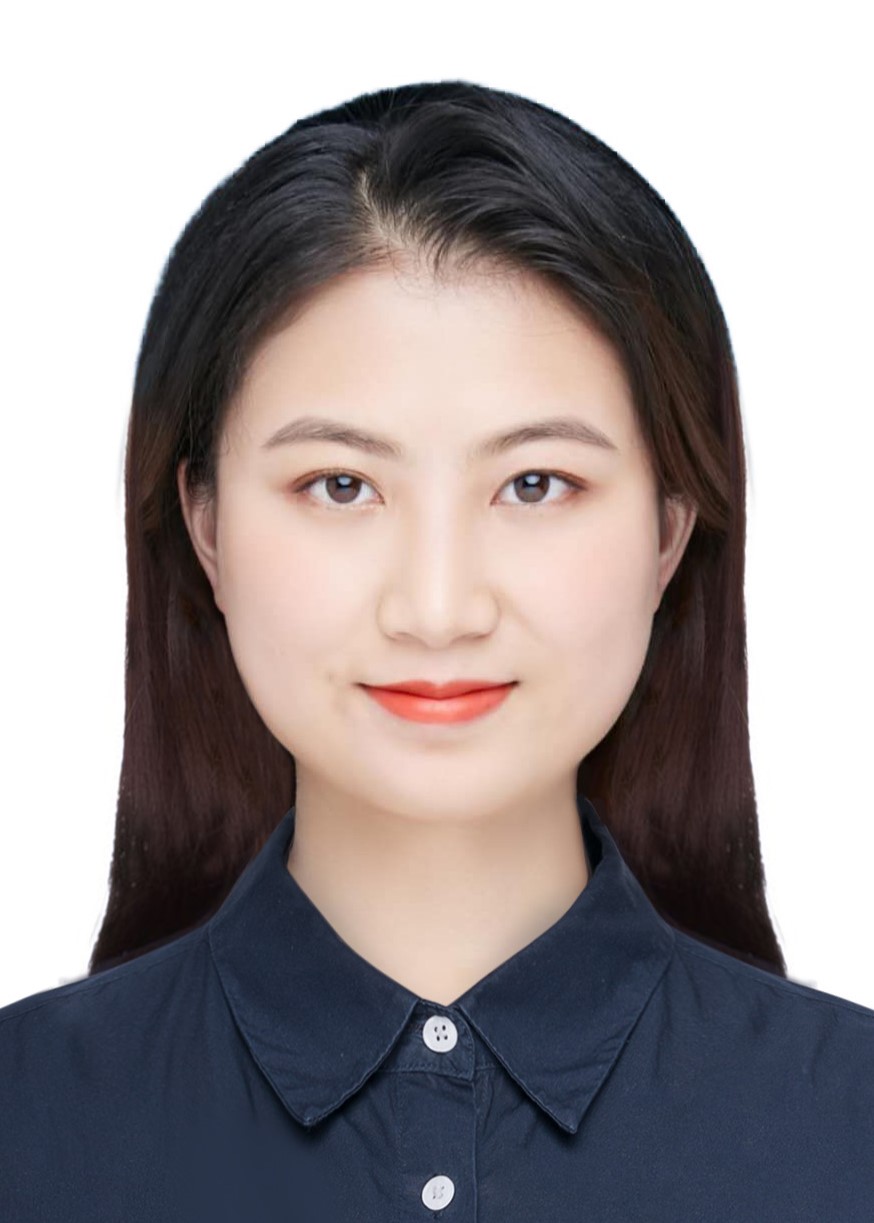}}]{Xie Yuhan}
received the B.Eng. degree in automation from the School of Automation Science and Electrical Engineering, Beihang University, in 2020. 
She is currently working toward an M.Phil. degree in robotics planning and control with the Department of Mechanical Engineering, University of Hong Kong.
Her research interests include unmanned aerial vehicles plan and control with deep reinforcement learning.
\end{IEEEbiography}

\begin{IEEEbiography}[{\includegraphics[width=1in,height=1.25in,clip,keepaspectratio]{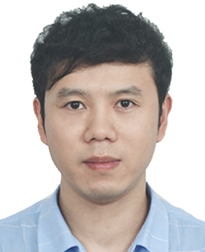}}]{Lu Peng}
obtained his BSc degree in automatic control and MSc degree in nonlinear flight control both from Northwestern Polytechnical University (NPU). He continued his journey on flight control at Delft University of Technology (TU Delft) where he received his PhD degree in 2016. After that, he shifted a bit from flight control and started to explore control for ground/construction robotics at ETH Zurich (ADRL lab) as a Postdoc researcher in 2016. He also had a short but nice journey at University of Zurich \& ETH Zurich (RPG group) where he was working on vision-based control for UAVs as a Postdoc researcher. He was an assistant professor in autonomous UAVs and robotics at Hong Kong Polytechnic University prior to joining the University of Hong Kong in 2020.

Prof. Lu has received several awards such as 3rd place in 2019 IROS autonomous drone racing competition and best graduate student paper finalist in AIAA GNC (top conference in aerospace). He serves as an associate editor for 2020 IROS (top conference in robotics) and session chair/co-chair for conferences like IROS and AIAA GNC for several times. He also gave a number of invited/keynote speeches at multiple conferences, universities and research institutes.
\end{IEEEbiography}

\end{document}